\pdfoutput=1

\documentclass[11pt]{article}

\usepackage[final]{acl}

\usepackage{times}
\usepackage{latexsym}

\usepackage[T1]{fontenc}

\usepackage[utf8]{inputenc}

\usepackage{microtype}

\usepackage{inconsolata}

\usepackage{graphicx}

\usepackage{booktabs}
\usepackage{CJKutf8}
\usepackage{multirow}
\usepackage{amsfonts}
\usepackage{amsmath}
\usepackage{xcolor, colortbl}
\usepackage{multirow, hhline}

\usepackage{enumitem}
\usepackage{float}

\definecolor{Gray}{gray}{0.90}
\newcolumntype{a}{>{\columncolor{Gray}}c}

\title{How Does Quantization Affect Multilingual LLMs?}

\author{
  \textbf{Kelly Marchisio\textsuperscript{1}},
  \textbf{Saurabh Dash\textsuperscript{2}},
  \textbf{Hongyu Chen\textsuperscript{1}},
  \textbf{Dennis Aumiller\textsuperscript{1}},
\\
  \textbf{Ahmet Üstün\textsuperscript{2}},
  \textbf{Sara Hooker\textsuperscript{2}},
  \textbf{Sebastian Ruder\textsuperscript{1}}
\\
\\
  \textsuperscript{1}Cohere
  \textsuperscript{2}Cohere For AI
\\
  \small{
    \textbf{Correspondence:} kelly@cohere.com
  }
}

\begin{document}
\maketitle
\begin{abstract}
Quantization techniques are widely used to improve inference speed and deployment of large language models. While a wide body of work examines the impact of quantization on LLMs in English, none have evaluated across languages. We conduct a thorough analysis of quantized multilingual LLMs, focusing on performance across languages and at varying scales. We use automatic benchmarks, LLM-as-a-Judge, and human evaluation, finding that (1) harmful effects of quantization are apparent in human evaluation, which automatic metrics severely underestimate: a 1.7\% average drop in Japanese across automatic tasks corresponds to a 16.0\% drop reported by human evaluators on realistic prompts; (2) languages are disparately affected by quantization, with non-Latin script languages impacted worst; and (3) challenging tasks like mathematical reasoning degrade fastest. As the ability to serve low-compute models is critical for wide global adoption of NLP technologies, our results urge consideration of multilingual performance as a key evaluation criterion for efficient models.
\end{abstract}

\section{Introduction}
\label{sec:intro}
Multilingual large language models (LLMs) have the power to bring modern language technology to the world, but only if they are cheap and reliable.  Known as the \textit{low-resource double bind}, underserved languages and severe compute constraints often geographically co-occur \citep{ahia2021lowresource}, meaning that for wide adoption, multilingual LLMs must be highly-performant \textit{and} lightweight.  

With the shift towards large models, quantization is a widely adopted technique to reduce cost, improve inference speed, and enable wider deployment of LLMs.  Work on quantization, however, is by-and-large evaluated in English only \citep[e.g.][]{xiao2023smoothquant, ahmadian-intruiging, frantar-gptq}. 
\begin{figure}[ht]
    \centering
    \includegraphics[width=1\linewidth]{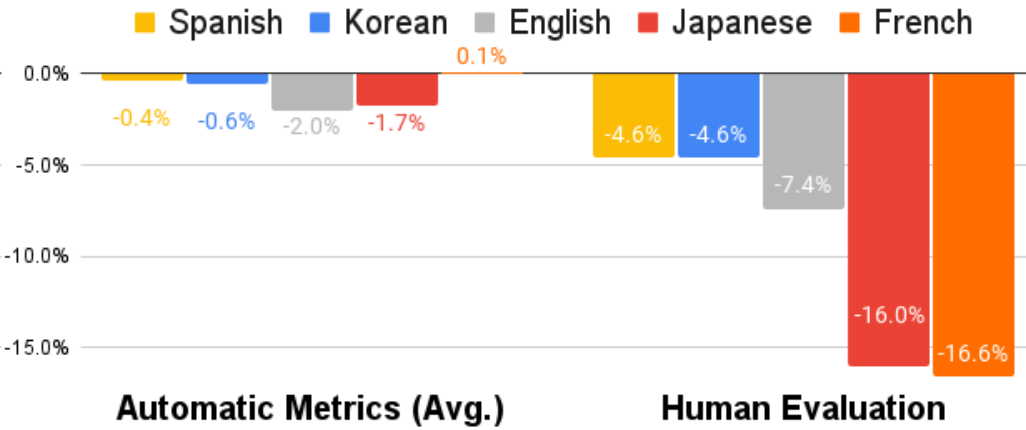}
    \caption{\textbf{Automatic metrics severely underestimate damage from quantization.} Shown: 103B W4 quantized Command model with group-wise scaling vs. FP16. Avg: mMMLU, FLORES, Language Confusion (LC). English avg: mMMLU, MGSM, monolingual LC.}
    \label{fig:auto-v-human-eval}
\end{figure}
No works to our knowledge have characterized the impact of quantization on the multilingual generation capabilities expected from modern LLMs. Ubiquitous use of compression techniques in the real world drives urgency to the question \textit{how are multilingual models impacted?}

Our question is timely, given recent work showing that compression techniques such as quantization and sparsity amplify disparate treatment of long-tail features, which may have implications for under-represented languages in multilingual LLMs \citep{Hooker2019WhatDC,hooker2020characterising,ahia2021lowresource,ogueji-etal-2022-intriguing}. 
Indeed, many model designs choices implicitly overfit to a handful of resource rich languages: from tokenizer choice, to weighting of training data, and to widely-used quantization techniques. 
Focusing on a small subset of high-resource languages in design degrades model performance for overlooked languages \citep{schwartz2022towards, Kotek2023GenderBA, Khandelwal2023CasteistBN, vashishtha2023evaluating,khondaker2023gptaraeval,pozzobon2024many}, introduces security vulnerabilities \citep{yong2023lowresource, nasr2023scalable, Li2023PrivacyIL, Lukas2023AnalyzingLO, deng2023multilingual}, and unfairly passes high costs to non-English users faced with high latency \citep{held2023material, durmus2023measuring,nicholas2023lost,ojo2023good,ahia2023languages}. 

We analyze four state-of-the-art (SOTA) multilingual LLMs across 3 different sizes ranging from 8 to 103 billion parameters and covering up to 23 languages, under various quantization techniques. Critically, it is vital that we move beyond automatic evaluation and gather \textit{real human feedback on performance cost}.  We thus perform multilingual human evaluation on challenging real-world prompts in addition to LLM-as-a-Judge and evaluation on standard automatic benchmarks such as multilingual MMLU \cite{hendrycks2020measuring}, MGSM \cite{shi2023language-mgsm}, and FLORES-200 \citep{nllb2022}. 
Across experimental set-ups we find that:

\setlist{nolistsep}
\begin{enumerate}[nolistsep]

    \item \textbf{Automatic metrics severely underestimate damage from quantization.} 
    Automatic evaluations estimate deterioration relative to FP16 across tasks at $-0.3\%$ (French) and $-1.7\%$ (Japanese) vs. $-16.6\%$ and $-16.0\%$ reported by human evaluators. See Figure \ref{fig:auto-v-human-eval}.\footnote{Figure excludes MGSM (not available for Korean.)} 
    
    \item \textbf{Quantization affects languages differently.} Degradation on automatic metrics appears negatively correlated with training data set size, and non-Latin script languages are more harmed on average. Across tasks, Latin-script languages scored $-0.7\%$ relative to FP16 for a 103B parameter model while non-Latin scripts scored $-1.9\%$.  For a smaller 8-billion parameter model, scores were $-3.0\%$ vs. $-3.7\%$.  

    \item\textbf{Challenging tasks degrade fastest.} Mathematical reasoning ($-13.1\%$), performance on real-world challenging prompts judged by humans ($-10.5\%$), and LLM-as-a-Judge ($-25.9\%$) results are severely degraded.

    \item \textbf{Occasionally, quantization brings benefits.} Similar to \citet{badshah2024quantifying}'s finding on English tasks, we find that quantization \textit{benefits} multilingual model performance in some cases: 
    e.g., an average 1.3\% boost across tasks for a 35B model quantized with W8A8. This aligns with findings on the benefit of other compression methods such as sparsity \citep{ahia2021lowresource, ogueji-etal-2022-intriguing}.
    
\end{enumerate}

As the first to broadly study the impact of quantization on multilingual LLMs, our work is part of a wider body of literature that considers the impact of model design choices on downstream performance. Our results urge attention to multilingual performance at all stages of system design.

\section{Background}
\label{sec:background}

Quantization compresses the weights and potentially activations of a neural network to lower-bit representations. Compression can be done by training the model at lower precision, known as Quantization Aware Training (QAT), or performed on the final model weights, known as Post Training Quantization (PTQ). Given the difficulties in training LLMs especially at precision lower than 16-bits floating point, PTQ methods which perform the quantization single-shot without needing gradient updates are highly desirable.
Training is completed at higher precision, then weights/activations are quantized without further training. In this work, we focus on \textit{post-training quantization} because of its simplicity and applicability at scale. PTQ of LLMs can be further categorized into:

\paragraph{\textbf{Weight-Only Quantization}} Weight matrices are quantized offline and the compressed matrices are loaded from memory during inference. Quantized weight matrices have a smaller memory footprint compared to FP16 ($2 \times$ smaller for 8-bit and almost $4 \times$ smaller for 4-bit), enabling inference with less compute. In memory-bound scenarios, it also enables faster inference due to fewer bytes transferred from GPU memory to the compute units. 

For a weight matrix $\mathbf{W} \in \mathbb{R}^{d_{in} \times d_{out}}$ and input $\mathbf{X} \in \mathbb{R}^{seq \times d_{in}}$, if only a single scaling factor is used for naive quantization (per-tensor), then the quantized weights are given by:
\begin{equation}
\mathbf{W}_Q = \left\lfloor \frac{\mathbf{W}}{\Delta} \right\rceil, \quad \Delta = \frac{\max(|\mathbf{W}|)}{2^{N-1} - 1}
\label{eq:naive-quant}
\end{equation}
where $\Delta \in \mathbb{R}$ denotes the scale, $N$ the bit precision, $|.|$ the absolute value over each element in $\mathbf{W}$ and $\left \lfloor . \right \rceil$ rounding to the nearest integer. When $\mathbf{W}_Q$ is used in a forward pass, it must be dequantized for multiplication with the higher-precision input matrix $\mathbf{X}$.  The result $\mathbf{Y}$ is $\mathbf{Y} = \mathbf{X}\Delta\mathbf{W}_Q$. Notably, the multiplication by $\Delta$ dequantizes $\mathbf{W}_Q$ (with error) so the result may be multiplied by the higher-precision $\mathbf{X}$.  $\mathbf{Y}$ has the same precision as $\mathbf{X}$.

A single scaling factor might not be enough if the distribution of parameters in the weight matrix has high variance; thus one could increase the granularity of quantization by using a scale for each output dimension (per-column), i.e., $\Delta \in \mathbb{R}^{d_{out}}$. However, when $N$ is aggressively lowered to 4 bits or lower, even per-column granularity might be insufficient to cover the range of values in a column. The granularity can be further increased by using a shared scale for a subset of input dimensions called groups ($g$), thus the scale $\Delta \in \mathbb{R}^{\frac{d_{in}}{g} \times d_{out}}$. A commonly used group size is 128.

Equation \ref{eq:naive-quant} gives the simplest way to quantize the weights. For $N \leq 4$ bits, using more advanced Weight-Only Quantization methods like GPTQ \citep{frantar-gptq} or AWQ \citep{lin2023awq} leads to better downstream performance.

\paragraph{\textbf{Weight-and-Activation Quantization}} As the name suggests, Weight-and-Activation Quantization quantizes the model activations alongside the weights. Unlike Weight-Only Quantization where weights can be quantized offline, quantization of activations happens at runtime. One could compute the quantization scales for various activations by using a small slice of training or validation data (static scaling) but this method typically has large degradation \citep{xiao2023smoothquant}. For minimal degradation, it is preferred to calculate the quantization scaling factor dynamically (dynamic scaling) for each input on-the-fly. While quantizing activations is more difficult, reducing the precision of the activations alongside the weights enables the usage of specialized low-precision matrix multiplication hardware in modern GPUs leading to up to $2 \times$ improvement in throughput. 
For a weight matrix $\mathbf{W} \in \mathbb{R}^{d_{in} \times d_{out}}$ and input $\mathbf{X} \in \mathbb{R}^{seq \times d_{in}}$, naive Weight-and-Activation Quantization with per-token input granularity and per-column weight granularity generates output $\mathbf{Y} \in \mathbb{R}^{seq \times d_{out}}$ by:
\begin{align}
{\mathbf{W}_Q}_{:,j} = \left\lfloor \frac{\mathbf{W}_{:,j}}{\Delta^{W}_{:,j}} \right\rceil, \Delta^{W}_{:,j} = \frac{\max(|\mathbf{W}_{:,j}|)}{2^{N-1} - 1} \\
{\mathbf{X}_Q}_{i,:} = \left\lfloor \frac{\mathbf{X}_{i,:}}{\Delta^{X}_{i,:}} \right\rceil, \Delta^{X}_{i,:} = \frac{\max(|\mathbf{X}_{i,:}|)}{2^{N-1} - 1}  
\end{align}
where $\Delta^{W} \in \mathbb{R}^{d_{out}}$ and $\Delta^{X} \in \mathbb{R}^{seq}$.  In the forward pass, $\mathbf{Y}$ is calculated as below, where $\odot$ denotes element-wise multiplication by broadcasting the elements to match the shape of the operands. The multiplication in lower-precision $\mathbf{X}_Q \mathbf{W}_Q$ is what leads to throughput gains. Multiplying by $\Delta^{W}$ and $\Delta^{X}$ de-quantizes the result so that $\mathbf{Y}$ has the same (higher) precision as the original $\mathbf{X}$.
\begin{align}
\mathbf{Y} = \Delta^X \odot (\mathbf{X}_Q \mathbf{W}_Q) \odot \Delta^W 
\end{align}

\section{Experimental Set-up}
\paragraph{Models} We use Command R+\footnote{\url{https://docs.cohere.com/docs/command-r-plus}}, 
Command R\footnote{\url{https://docs.cohere.com/docs/command-r}}, and Aya 23 models \citep{aryabumi2024aya} as representative of SOTA multilingual LLMs. Command models are 103 and 35 billion parameters (R+/R). Aya 23 models are 35 and 8 billion parameters. We quantize the weights available on HuggingFace.


\paragraph{Quantization} 
For Command R/R+ (35B/103B), we evaluate \textbf{weight-only quantization} at 8-bit (\textbf{W8} with per-column scaling) and 4-bit (\textbf{W4-g} with group-wise scaling using GPTQ \citep{frantar-gptq}), as well as \textbf{weight-and-activation quantization} at 8-bit (\textbf{W8A8} with per-column scaling for weights and per-token scaling for activations). 

When trained with the right hyper-parameters, naive Weight-and-Activation Quantization has minimal degradation \citep{ahmadian-intruiging}. Otherwise, SmoothQuant \citep{xiao2023smoothquant} may smoothen the activation distributions to be more amenable to quantization. We explore \textbf{W8A8-SmoothQuant} (W8A8 with SmoothQuant) for Command R+ (103B) and a 4-bit weight-only quantized variant with column-wise scaling (\textbf{W4}) to understand the impact of scaling granularity at extremely low-bit precision. We use 128 English samples for calibration for SmoothQuant and GPTQ 
\citep{frantar-gptq,xiao2023smoothquant}.

For Aya 23 8B and 35B, we use bitsandbytes\footnote{\url{https://github.com/TimDettmers/bitsandbytes}} for 8-bit and 4-bit quantization. This uses LLM.int8() \citep{dettmers2022llm}---similar to W8A8 except it performs some computations in FP16. The 4-bit uses the NF4 datatype \citep{dettmers2023qlora} to perform Quantile Quantization which limits degradation at the expense of inference speedups.

\subsection{Automatic Evaluation}
\label{sec:eval-auto}
We evaluate in 10 primary languages: \textit{Arabic}, \textit{French}, \textit{German}, \textit{English}, \textit{Spanish}, \textit{Italian}, \textit{Portuguese}, \textit{Korean}, \textit{Japanese}, and \textit{Chinese}. Quantized models are compared to the original \textbf{FP16} versions, and we primarily report results as \textbf{relative degradation} compared to this FP16 baseline:
\begin{equation}
\%\Delta = \frac{\text{score}_\text{quantized} - \text{score}_\text{FP16}}{\text{score}_\text{FP16}}*100
\end{equation}
Raw numeric results are in the Appendix.  Results are averaged over 5 runs.\footnote{k=0, p=0.75, temp=0.3, except mMMLU, which, as a QA eval, is run deterministically with temp=0.} 

\paragraph{Multilingual MMLU} 14,000+ multi-domain multiple-choice questions.  We translate MMLU \citep{hendrycks2020measuring} to 9 languages with Google Translate and call it \textbf{mMMLU}. We measure 5-shot accuracy. Example in Table \ref{tab:app-mmmlu-prompt}.

\paragraph{MGSM \citep{shi2023language-mgsm}} Generative mathematics test set manually translated from GSM8K \citep{cobbe2021training}. Of our target languages, it is available for German, Spanish, French, Japanese, Chinese.  We report accuracy for each language.

\paragraph{FLORES-200 \citep{costa2022no}} This well-known multi-way parallel test set evaluates translation capabilities.  We translate into and out of English, and report SacreBLEU \cite{post-2018-call}.

\paragraph{Language Confusion \citep{marchisio2024understandingmitigatinglanguageconfusion}} These test sets assess a model's ability to respond in a user's desired language.  In the monolingual setting, prompts are in language $l$ and the model must respond in language $l$. For instance, a user prompts in Arabic, so implicitly desires an Arabic response. In the cross-lingual variant, a prompt is provided in English but the user requests output in a different language $l'$.\footnote{An example from the \textit{Okapi} subsection of the evaluation is: ``Reply in Spanish. Explain a common misconception about your topic. Topic: Using AI to Augment Human Capabilities''} \textit{fastText} \cite{joulin2016bag} language identification is run over the output.  We report \textit{line-level pass rate} (\textbf{LPR)}, i.e., the percentage of responses for which all lines in the response are in the user's desired language.

\paragraph{Aya Evaluation}
Aya 23 models are evaluated using an extended version of the Aya evaluation setup \citep{aryabumi2024aya} using the unseen discriminative tasks---those where there is no dataset in the models’ training mixture from the same task categories (XWinograd \citep{muennighoff2022crosslingual}, XCOPA \citep{ponti2020xcopa}, XStoryCloze \citep{lin-etal-2022-shot}), mMMLU \citep[Okapi;][]{dac2023okapi}, MGSM, and Belebele \citep{bandarkar2023belebele} from \texttt{eval-harness} \citep{eval-harness}.\footnote{We follow \citet{ustun2024aya}: each evaluation is run once; For FLORES, no sampling is used and metric is spBLEU.} We evaluate models on languages included in the covered 23 languages, except for the unseen tasks where we use all available languages.\footnote{mMMLU: ar, de, es, fr, hi, id, it, nl, pt, ro, ru, uk, vi, zh. MGSM: de, es, fr, ja, ru, zh. Belebele: \{mMMLU\} + cs, fa, el, ja, ko, pl, tr. FLORES: \{Belebele\} + he.} Aya evaluations allow us to add: \textit{Czech, Greek, Hebrew, Hindi, Indonesian, Dutch, Persian, Polish, Romanian, Russian, Turkish, Ukrainian, Vietnamese}.

\subsection{Human Evaluation}
\label{sec:eval-human}
We run human evaluation in \textit{Spanish}, \textit{French}, \textit{Korean}, \textit{Japanese}, and \textit{English}. 

\paragraph{Internal Evaluation Suite} 150 diverse prompts designed to be more complex than public evaluation benchmarks. As such, we expect greater degradation with increased quantization given the difficulty of the samples. Prompts for all four languages are translated by humans from an English seed prompt, ensuring that respective language-specific subsets share the same prompts.

\paragraph{Aya Dolly-200 ~\citep{ayadata2024}}
We use multilingual data from the Aya Evaluation Suite to assess open-ended generation capabilities. For Korean and Japanese, we use prompts from the Aya Dolly-200 test set (\textbf{dolly-machine-translated}), which are automatically translated from English Dolly-15k \citep{DatabricksBlog2023DollyV2} then human-curated to avoid references requiring specific cultural or geographic knowledge.  
For French and Spanish, we use \textbf{dolly-human-edited}, a human post-edited version of \textbf{dolly-machine-translated}. For each language, we evaluate using the first 150 prompts. 

\paragraph{Annotation}
Annotations and translations were completed by native-level speakers of the respective languages who are also fluent in English.\footnote{Paid hourly, above min. wage of country of employment.}
The annotation interface supports pairwise evaluation. Annotators see a prompt and two (shuffled) completions of the FP16 model and a quantized variant which they rate on a 5-point Likert scale, then express a preference (tie, weak preference, strong preference). We encourage annotators to avoid ties. Win rates are based on ranking preferences alone.

\begin{table*}[htb]
\resizebox{\textwidth}{!}{%
\begin{tabular}{ll|c|rr|rr|rr|rr|rr|rr}
\toprule
 &    & \multicolumn{1}{c|}{\textbf{Avg.}} &    &    &    &   & \multicolumn{4}{c|}{\textbf{FLORES}} &   \multicolumn{4}{c}{\textbf{Language Confusion}}    \\
 &
   &
  \multicolumn{1}{c|}{\textbf{Rel. $\%\Delta$}} &
  \multicolumn{2}{c|}{\textbf{mMMLU}} &
  \multicolumn{2}{c|}{\textbf{MGSM}} &
  \multicolumn{2}{c|}{\textbf{En$\rightarrow$L2}} &
  \multicolumn{2}{c|}{\textbf{L2$\rightarrow$En}} &
  \multicolumn{2}{c|}{\textbf{Monolingual}} &
  \multicolumn{2}{c}{\textbf{Cross-lingual}} \\  \midrule
 & FP16 & \multicolumn{1}{c|}{-} & 66.7 & \multicolumn{1}{c|}{-} & 70.6 & \multicolumn{1}{c|}{-} & 37.7 & \multicolumn{1}{c|}{-} & 39.6 & \multicolumn{1}{c|}{-} & 99.2 & \multicolumn{1}{c|}{-} & 91.5 & \multicolumn{1}{c}{-} \\
 & W8 & \cellcolor[HTML]{FDF9F9}-0.2\% & 66.7 & \cellcolor[HTML]{FFFFFF}0.0\% & 69.9 & \cellcolor[HTML]{FAE6E4}-1.0\% & 37.7 & \cellcolor[HTML]{FFFFFF}0.0\% & 39.6 & \cellcolor[HTML]{FEFEFE}0.0\% & 99.2 & \cellcolor[HTML]{FFFFFF}0.0\% & 91.2 & \cellcolor[HTML]{FDF7F7}-0.3\% \\ \hhline{~*{14}{-}}
 & W8A8-sq & \cellcolor[HTML]{FCF1F0}-0.5\% & 66.3 & \cellcolor[HTML]{FCF1F0}-0.5\% & 69.5 & \cellcolor[HTML]{F6D4D1}-1.6\% & 37.8 & \cellcolor[HTML]{FBFEFC}0.2\% & 39.1 & \cellcolor[HTML]{F8DDDB}-1.3\% & 99.2 & \cellcolor[HTML]{FFFFFF}0.0\% & 91.5 & \cellcolor[HTML]{FEFFFF}0.1\% \\
 & W8A8 & \cellcolor[HTML]{FAE9E7}-0.8\% & 65.6 & \cellcolor[HTML]{F6D3D0}-1.7\% & 69.8 & \cellcolor[HTML]{F9E2E0}-1.1\% & 37.7 & \cellcolor[HTML]{FFFFFF}0.0\% & 39.1 & \cellcolor[HTML]{F8DEDC}-1.2\% & 99.4 & \cellcolor[HTML]{FBFEFC}0.2\% & 90.4 & \cellcolor[HTML]{F8DFDD}-1.2\% \\ \hhline{~*{14}{-}}
 & W4-g & \cellcolor[HTML]{FAE6E5}-0.9\% & 65.7 & \cellcolor[HTML]{F7D9D6}-1.4\% & 68.6 & \cellcolor[HTML]{F0B3AD}-2.9\% & 37.8 & \cellcolor[HTML]{F6FCF9}0.4\% & 39.4 & \cellcolor[HTML]{FCF2F1}-0.5\% & 99.2 & \cellcolor[HTML]{FFFFFF}0.0\% & 90.5 & \cellcolor[HTML]{F9E2E0}-1.1\% \\
\multirow{-6}{*}{103B} & W4 & \cellcolor[HTML]{F2BEB9}-2.5\% & 63.8 & \cellcolor[HTML]{E98D85}-4.3\% & 64.4 & \cellcolor[HTML]{E67C73}-8.8\% & 37.1 & \cellcolor[HTML]{F6D3D0}-1.6\% & 39.0 & \cellcolor[HTML]{F7D5D2}-1.6\% & 99.3 & \cellcolor[HTML]{FCFEFD}0.1\% & 92.8 & \cellcolor[HTML]{DBF1E6}1.4\% \\ \midrule\midrule
 & FP16 & \multicolumn{1}{c|}{-} & 59.4 & \multicolumn{1}{c|}{-} & 49.8 & \multicolumn{1}{c|}{-} & 32.4 & \multicolumn{1}{c|}{-} & 35.5 & \multicolumn{1}{c|}{-} & 98.7 & \multicolumn{1}{c|}{-} & 66.5 & \multicolumn{1}{c}{-} \\
 & W8 & \cellcolor[HTML]{FDF9F9}-0.2\% & 59.3 & \cellcolor[HTML]{FEFDFD}-0.1\% & 49.4 & \cellcolor[HTML]{FBECEB}-0.7\% & 32.3 & \cellcolor[HTML]{FEFBFA}-0.2\% & 35.4 & \cellcolor[HTML]{FEF9F9}-0.2\% & 98.8 & \cellcolor[HTML]{FDFFFE}0.1\% & 66.3 & \cellcolor[HTML]{FDF9F9}-0.2\% \\
 & W8A8 & \cellcolor[HTML]{FCFEFD}0.2\% & 59.3 & \cellcolor[HTML]{FEFAFA}-0.2\% & 47.1 & \cellcolor[HTML]{E67C73}-5.5\% & 32.9 & \cellcolor[HTML]{D6EFE3}1.6\% & 35.8 & \cellcolor[HTML]{E8F6EF}0.9\% & 99.0 & \cellcolor[HTML]{F7FCF9}0.3\% & 68.9 & \cellcolor[HTML]{A0D9BD}3.7\% \\
\multirow{-4}{*}{35B} & W4-g & \cellcolor[HTML]{F0B4AF}-2.8\% & 58.2 & \cellcolor[HTML]{F5CBC7}-2.0\% & 43.3 & \cellcolor[HTML]{E67C73}-13.1\% & 31.7 & \cellcolor[HTML]{F5CDC9}-1.9\% & 35.3 & \cellcolor[HTML]{FBEDEB}-0.7\% & 98.3 & \cellcolor[HTML]{FCF3F3}-0.4\% & 67.1 & \cellcolor[HTML]{E7F5EE}1.0\% \\ \bottomrule
\end{tabular}%
}
\caption{\textbf{Per-dataset average performance across non-English languages for 103B and 35B Command models at varying levels of quantization.}  \%$\Delta$ the relative performance vs. FP16 [ex., for MGSM at W4-g on the 35B: $\frac{43.3 - 49.8}{49.8}*100 = -13.1\%$.] Languages: ar, de, es, fr, it, ja, ko, pt, zh; except MGSM: de, es, fr, ja, zh.  
Any discrepancy is due to rounding: raw scores and \textbf{\%$\Delta$} were calculated at full precision. 
} 
\label{tab:main-overall}
\end{table*}

\begin{table*}[htb]
\centering
\resizebox{\textwidth}{!}{%
\begin{tabular}{ll|c|rr|rr|rr|rr|rr|rr}
\toprule
 &    & \multicolumn{1}{c|}{\textbf{Avg.}} &    &    &    &   & \multicolumn{4}{c|}{\textbf{FLORES}} & & & \multicolumn{2}{c}{\textbf{Unseen}}    \\
 &
   &
  \multicolumn{1}{c|}{\textbf{Rel. $\%\Delta$}} &
  \multicolumn{2}{c|}{\textbf{mMMLU}} &
  \multicolumn{2}{c|}{\textbf{MGSM}} &
  \multicolumn{2}{c|}{\textbf{En$\rightarrow$L2}} &
  \multicolumn{2}{c|}{\textbf{L2$\rightarrow$En}} &
  \multicolumn{2}{c|}{\textbf{Belebele}} &
  \multicolumn{2}{c}{\textbf{Tasks}} \\ \midrule
  \multicolumn{1}{c}{} &
  FP16 &
  - &
  58.2 &
  \multicolumn{1}{c|}{-} &
  51.2 &
  \multicolumn{1}{c|}{-} &
  37.8 &
  \multicolumn{1}{c|}{-} &
  42.9 &
  \multicolumn{1}{c|}{-} &
  77.6 &
  \multicolumn{1}{c|}{-} &
  70.8 &
  \multicolumn{1}{c}{-} \\
\multicolumn{1}{c}{} &
  W8 &
  \cellcolor[HTML]{FEFFFE}0.1\% &
  57.9 &
  \cellcolor[HTML]{FCF2F1}-0.5\% &
  52.1 &
  \cellcolor[HTML]{D0ECDF}1.8\% &
  37.9 &
  \cellcolor[HTML]{F9FDFB}0.3\% &
  43.0 &
  \cellcolor[HTML]{FEFFFE}0.1\% &
  77.1 &
  \cellcolor[HTML]{FBEEED}-0.6\% &
  70.6 &
  \cellcolor[HTML]{FDF9F8}-0.2\% \\
\multicolumn{1}{c}{\multirow{-3}{*}{Aya 35B}} &
  W4 &
  \cellcolor[HTML]{F1B7B3}-2.9\% &
  56.6 &
  \cellcolor[HTML]{F1B7B3}-2.7\% &
  48.1 &
  \cellcolor[HTML]{E67C73}-6.0\% &
  37.2 &
  \cellcolor[HTML]{F8DBD9}-1.4\% &
  42.4 &
  \cellcolor[HTML]{F8DFDC}-1.2\% &
  73.0 &
  \cellcolor[HTML]{E67C73}-5.9\% &
  70.5 &
  \cellcolor[HTML]{FDF6F5}-0.3\% \\ \midrule\midrule
 &
  FP16 &
  - &
  48.2 &
  \multicolumn{1}{c|}{-} &
  34.7 &
  \multicolumn{1}{c|}{-} &
  34.8 &
  \multicolumn{1}{c|}{-} &
  39.5 &
  \multicolumn{1}{c|}{-} &
  64.8 &
  \multicolumn{1}{c|}{-} &
  67.6 &
  \multicolumn{1}{c}{-} \\
 &
  W8 &
  \cellcolor[HTML]{F7FCFA}0.3\% &
  47.8 &
  \cellcolor[HTML]{FAE7E6}-0.9\% &
  35.4 &
  \cellcolor[HTML]{C9E9D9}2.1\% &
  34.8 &
  \cellcolor[HTML]{FAFDFC}0.2\% &
  39.7 &
  \cellcolor[HTML]{F4FBF7}0.5\% &
  64.6 &
  \cellcolor[HTML]{FDF6F6}-0.3\% &
  67.6 &
  \cellcolor[HTML]{FEFFFE}0.1\% \\
\multirow{-3}{*}{Aya 8B} &
  W4 &
  \cellcolor[HTML]{EC9E97}-3.7\% &
  46.7 &
  \cellcolor[HTML]{EEABA5}-3.2\% &
  32.1 &
  \cellcolor[HTML]{E67C73}-7.5\% &
  34.1 &
  \cellcolor[HTML]{F5CFCC}-1.8\% &
  39.1 &
  \cellcolor[HTML]{FAE5E3}-1.0\% &
  59.3 &
  \cellcolor[HTML]{E67C73}-8.5\% &
  67.5 &
  \cellcolor[HTML]{FEFAFA}-0.2\%

\\ \bottomrule
\end{tabular}%
 }
\caption{\textbf{Per-dataset average performance across non-English languages for 35B and 8B Aya 23 models at varying levels of quantization.} \%$\Delta$ is relative performance vs. FP16. We follow the evaluation setup of \citet{aryabumi2024aya} and evaluate on languages in the 23 languages list.  On ``Unseen Tasks'' (XWinograd, XCOPA, XStoryCloze), we use all the available languages. 
See Section \ref{sec:eval-auto} for details and language list.  
} 
\label{tab:aya-overall}
\end{table*}

\subsection{LLM/RM-as-a-Judge}  
Because human evaluation is costly and time-intensive, it is common to use an ``LLM-as-a-Judge'' to rate model completions \citep[e.g.][]{alpaca_eval, zheng2023judging}.  Reward models (RMs) can also simulate human preference.  A RM scores multiple completions given the same prompt, and the prompt-completion pair with the higher score is deemed preferred.  We call this \emph{RM-as-a-Judge}. 

We assess quantized model outputs using LLM- and RM-as-a-Judge.  In the former, an LLM selects a preferred response from a 
\texttt{$<$instruction, modelA\_completion, modelB\_completion$>$} tuple (see Table \ref{tab:app-llm-as-a-judge-prompt-gpt4}).  We use GPT-4\footnote{\texttt{turbo (gpt-4-1106-preview)}: \url{https://platform.openai.com/docs/models/gpt-4-turbo-and-gpt-4}} as an LLM proxy judge following \citet{ustun2024aya} and \citet{aryabumi2024aya}. We randomize the order of model outputs to minimize bias.  For RM-as-a-Judge, a multilingual RM scores \texttt{$<$prompt, completion$>$} 
pairs for each model output, over which we calculate win-rate. We report win-rates of quantized models versus the FP16 baseline. 

We assess the outputs of quantized models over the \textit{Internal Evaluation Suite} and \textit{Aya Dolly-200} described in Section \ref{sec:eval-human}.  We use the same prompt and completion pairs as in human evaluation, which provides the ability to relate LLM/RM-as-a-Judge performance with human evaluation.

\section{Results}

To clearly see the many-faceted impact of quantization, we discuss our results by quantization level (\S\ref{sec:results-quantization-level}), by task (\S\ref{sec:results-task}), by language (\S\ref{sec:specific-lan}), by model size (\S\ref{sec:results-model-size}), and by quantization strategy (\S\ref{sec:strategies}). We then report LLM-as-a-Judge and RM-as-a-Judge (\S\ref{sec:results-llm-as-a-judge}) and human evaluation results (\S\ref{sec:results-human-evaluation}).

\subsection{By Quantization Level} \label{sec:results-quantization-level}
\textit{How do different levels of quantization affect downstream performance?}

\paragraph{Command R (35B) and R+ (103B)}   
In Table \ref{tab:main-overall}, we aggregate results of each metric for each level of quantization. We average scores across languages, then calculate the relative percentage drop from FP16.\footnote{Ex. For 103B \textbf{W4-g} MGSM, scores were: \{de: 71.2, es: 75.7, fr: 69.0, ja: 58.0, zh: 68.9\}, thus the average score was 68.6---a 2.9\% drop from FP16 ($\frac{68.6 - 70.6}{70.6} = -0.029$).}  We discuss results of \textbf{W8}, \textbf{W8A8}, and \textbf{W4-g} quantization, which are variants available for both Command model sizes. Most results follow intuition: greater quantization leads to larger performance degradation: $-0.2\%$ for \textbf{W8}, $-0.8\%$ for \textbf{W8A8}, and $-0.9\%$ for \textbf{W4-g} of the 103B model.  An exception is \textbf{W8A8} for the 35B which shows a slight boost overall due to higher performance on translation and language confusion evaluations.  

\begin{table*}[htb]
\centering
\resizebox{\textwidth}{!}{%
\begin{tabular}{@{}llrrrrrrrrr||rrr}
\toprule
                       &                           & \multicolumn{1}{c}{\textbf{ar}} & \multicolumn{1}{c}{\textbf{de}} & \multicolumn{1}{c}{\textbf{es}} & \multicolumn{1}{c}{\textbf{fr}} & \multicolumn{1}{c}{\textbf{it}} & \multicolumn{1}{c}{\textbf{ja}} & \multicolumn{1}{c}{\textbf{ko}} & \multicolumn{1}{c}{\textbf{pt}} & \multicolumn{1}{c||}{\textbf{zh}} & \multicolumn{1}{c}{\textbf{Avg}} & \multicolumn{1}{c}{\textbf{Ltn/IE}} & \multicolumn{1}{c}{\textbf{$\neg$}} \\ \midrule
                       & \multicolumn{1}{l|}{W8}   & \cellcolor[HTML]{FEFEFE}0.0\%   & \cellcolor[HTML]{FDFEFE}0.1\%   & \cellcolor[HTML]{FEFFFF}0.0\%   & \cellcolor[HTML]{FEFDFD}0.0\%   & \cellcolor[HTML]{FEFEFE}0.0\%   & \cellcolor[HTML]{FEFFFE}0.1\%   & \cellcolor[HTML]{FEFEFE}0.0\%   & \cellcolor[HTML]{FCF4F3}-0.4\%  & \cellcolor[HTML]{FDF8F8}-0.2\%   & \cellcolor[HTML]{FEFDFD}-0.1\%    & \cellcolor[HTML]{FEFDFD}-0.1\%      & \cellcolor[HTML]{FEFDFD}-0.1\%   \\ \hhline{~*{13}{-}}
                       & \multicolumn{1}{l|}{W8A8-sq} & \cellcolor[HTML]{FCF0EF}-0.6\%  & \cellcolor[HTML]{FAFDFC}0.2\%   & \cellcolor[HTML]{FDF6F6}-0.3\%  & \cellcolor[HTML]{FCFEFD}0.1\%   & \cellcolor[HTML]{FBEEEC}-0.6\%  & \cellcolor[HTML]{FDF6F5}-0.3\%  & \cellcolor[HTML]{FEFBFB}-0.1\%  & \cellcolor[HTML]{FBEDEC}-0.7\%  & \cellcolor[HTML]{FBEAE9}-0.8\%  & \cellcolor[HTML]{FDF6F5}-0.3\%   & \cellcolor[HTML]{FDF8F7}-0.3\%      & \cellcolor[HTML]{FCF3F2}-0.4\%        \\
                       & \multicolumn{1}{l|}{W8A8} & \cellcolor[HTML]{F8DDDB}-1.3\%  & \cellcolor[HTML]{FAE7E6}-0.9\%  & \cellcolor[HTML]{FCF3F2}-0.5\%  & \cellcolor[HTML]{FCF3F2}-0.5\%  & \cellcolor[HTML]{FAE9E7}-0.8\%  & \cellcolor[HTML]{FDF7F6}-0.3\%  & \cellcolor[HTML]{F8DCD9}-1.3\%  & \cellcolor[HTML]{FBEBE9}-0.8\%  & \cellcolor[HTML]{FAE7E5}-0.9\%   & \cellcolor[HTML]{FBEAE8}-0.8\%    & \cellcolor[HTML]{FBEDEC}-0.7\%      & \cellcolor[HTML]{FAE6E4}-0.9\%   \\ \hhline{~*{13}{-}}
                        & \multicolumn{1}{l|}{W4-g}  & \cellcolor[HTML]{FAE9E7}-0.8\%  & \cellcolor[HTML]{FEFAFA}-0.2\%  & \cellcolor[HTML]{FDF5F4}-0.4\%  & \cellcolor[HTML]{FEFFFE}0.1\%   & \cellcolor[HTML]{FDF5F5}-0.4\%  & \cellcolor[HTML]{FCF3F3}-0.4\%  & \cellcolor[HTML]{FCF0EF}-0.6\%  & \cellcolor[HTML]{F8DFDD}-1.2\%  & \cellcolor[HTML]{FAE6E5}-0.9\%   & \cellcolor[HTML]{FCF1F0}-0.5\%    & \cellcolor[HTML]{FCF4F3}-0.4\%      & \cellcolor[HTML]{FBEDEB}-0.7\%   \\ 
                        \multirow{-5}{*}{103B} & \multicolumn{1}{l|}{W4}      & \cellcolor[HTML]{F9E4E2}-1.0\%  & \cellcolor[HTML]{FBEFED}-0.6\%  & \cellcolor[HTML]{FDFEFE}0.1\%   & \cellcolor[HTML]{FAE9E7}-0.8\%  & \cellcolor[HTML]{F9E0DD}-1.2\%  & \cellcolor[HTML]{F7D9D7}-1.4\%  & \cellcolor[HTML]{F0B2AC}-2.9\%  & \cellcolor[HTML]{FBEAE9}-0.8\%  & \cellcolor[HTML]{F3C2BE}-2.3\%  & \cellcolor[HTML]{F8DFDC}-1.2\%   & \cellcolor[HTML]{FBEDEC}-0.7\%      & \cellcolor[HTML]{F5CCC9}-1.9\%        \\ \midrule\midrule
                       & \multicolumn{1}{l|}{W8}   & \cellcolor[HTML]{F9FDFB}0.3\%   & \cellcolor[HTML]{FCF2F1}-0.5\%  & \cellcolor[HTML]{FEFCFB}-0.1\%  & \cellcolor[HTML]{FEFAF9}-0.2\%  & \cellcolor[HTML]{FDF5F4}-0.4\%  & \cellcolor[HTML]{F7FCFA}0.3\%   & \cellcolor[HTML]{FEFBFA}-0.1\%  & \cellcolor[HTML]{FEFFFE}0.1\%   & \cellcolor[HTML]{FDF8F7}-0.3\%   & \cellcolor[HTML]{FEFCFC}-0.1\%    & \cellcolor[HTML]{FDF9F8}-0.2\%      & \cellcolor[HTML]{FEFFFF}0.0\%    \\
                       & \multicolumn{1}{l|}{W8A8} & \cellcolor[HTML]{CBEADB}2.0\%   & \cellcolor[HTML]{C0E6D3}2.5\%   & \cellcolor[HTML]{EFF9F4}0.7\%   & \cellcolor[HTML]{E5F5ED}1.0\%   & \cellcolor[HTML]{E1F3EA}1.2\%   & \cellcolor[HTML]{E2F4EB}1.1\%   & \cellcolor[HTML]{E9F6F0}0.9\%   & \cellcolor[HTML]{DAF0E6}1.4\%   & \cellcolor[HTML]{E4F5ED}1.0\%    & \cellcolor[HTML]{DEF2E8}1.3\%     & \cellcolor[HTML]{DDF1E7}1.3\%       & \cellcolor[HTML]{DFF2E9}1.3\%    \\
\multirow{-3}{*}{35B}  & \multicolumn{1}{l|}{W4-g}  & \cellcolor[HTML]{F9E1DF}-1.1\%  & \cellcolor[HTML]{F9E2E0}-1.1\%  & \cellcolor[HTML]{FDFFFE}0.1\%   & \cellcolor[HTML]{FDF7F6}-0.3\%  & \cellcolor[HTML]{FEFDFC}-0.1\%  & \cellcolor[HTML]{F3C3BF}-2.3\%  & \cellcolor[HTML]{F7D9D7}-1.4\%  & \cellcolor[HTML]{FBEFEE}-0.6\%  & \cellcolor[HTML]{F8DBD9}-1.3\%   & \cellcolor[HTML]{FAE7E5}-0.9\%    & \cellcolor[HTML]{FDF4F3}-0.4\%      & \cellcolor[HTML]{F7D6D4}-1.5\%   \\ \bottomrule
\end{tabular}%
}
\caption{\textbf{Per-language relative performance (\%$\Delta$) vs. FP16, averaged over mMMLU, FLORES, and Language Confusion tasks.} Ltn/IE are Latin-script/Indo-European languages: de, es, fr, it, pt.  $\neg$ are the rest: ar, ja, ko, zh.}
\label{tab:avgs-no-mgsm}
\end{table*}

\begin{table}[htb]
\centering
\resizebox{\columnwidth}{!}{%
\begin{tabular}{@{}ll|rrrrr||rrr@{}}
\toprule
                       &         & \multicolumn{1}{c}{\textbf{de}} & \multicolumn{1}{c}{\textbf{es}} & \multicolumn{1}{c}{\textbf{fr}} & \multicolumn{1}{c}{\textbf{ja}} & \multicolumn{1}{c||}{\textbf{zh}} & \multicolumn{1}{c|}{\textbf{Avg}} & \multicolumn{1}{c}{\textbf{Ltn/IE}} & \multicolumn{1}{c}{\textbf{$\neg$}} \\ \midrule
                       & W8      & \cellcolor[HTML]{FCFEFD}0.1\%   & \cellcolor[HTML]{FEFBFB}-0.1\%  & \cellcolor[HTML]{FDF7F6}-0.3\%  & \cellcolor[HTML]{FDF4F4}-0.4\%  & \cellcolor[HTML]{FDF9F8}-0.2\%   & \cellcolor[HTML]{FEFAFA}-0.2\%    & \cellcolor[HTML]{FEFCFC}-0.1\%      & \cellcolor[HTML]{FDF7F6}-0.3\%        \\ \hhline{~*{9}{-}}
                       & W8A8-sq & \cellcolor[HTML]{F6FCF9}0.4\%   & \cellcolor[HTML]{FAE8E6}-0.9\%  & \cellcolor[HTML]{FEFBFB}-0.1\%  & \cellcolor[HTML]{FDF7F6}-0.3\%  & \cellcolor[HTML]{F8DEDC}-1.2\%   & \cellcolor[HTML]{FCF3F2}-0.4\%    & \cellcolor[HTML]{FDF9F9}-0.2\%      & \cellcolor[HTML]{FBEAE9}-0.8\%        \\
                       & W8A8    & \cellcolor[HTML]{FDF4F4}-0.4\%  & \cellcolor[HTML]{FAE5E3}-1.0\%  & \cellcolor[HTML]{FCF0EF}-0.6\%  & \cellcolor[HTML]{FEFBFB}-0.1\%  & \cellcolor[HTML]{F8DDDB}-1.3\%   & \cellcolor[HTML]{FBEDEC}-0.7\%    & \cellcolor[HTML]{FBEEED}-0.6\%      & \cellcolor[HTML]{FBECEB}-0.7\%        \\ \hhline{~*{9}{-}}
                       & W4-g     & \cellcolor[HTML]{FCF3F2}-0.5\%  & \cellcolor[HTML]{FCF1F0}-0.5\%  & \cellcolor[HTML]{FDF6F5}-0.3\%  & \cellcolor[HTML]{F6D3D0}-1.7\%  & \cellcolor[HTML]{F9E2E0}-1.1\%   & \cellcolor[HTML]{FAE9E8}-0.8\%    & \cellcolor[HTML]{FCF3F2}-0.4\%      & \cellcolor[HTML]{F8DAD8}-1.4\%        \\
\multirow{-5}{*}{103B} & W4      & \cellcolor[HTML]{F3C1BD}-2.3\%  & \cellcolor[HTML]{F9E2E0}-1.1\%  & \cellcolor[HTML]{F6D3D0}-1.7\%  & \cellcolor[HTML]{EFB0AA}-3.0\%  & \cellcolor[HTML]{EDA29C}-3.5\%   & \cellcolor[HTML]{F3C2BD}-2.3\%    & \cellcolor[HTML]{F6D2CF}-1.7\%      & \cellcolor[HTML]{EEA9A3}-3.3\%        \\ \midrule
                       & W8      & \cellcolor[HTML]{FBEFED}-0.6\%  & \cellcolor[HTML]{FDF7F7}-0.3\%  & \cellcolor[HTML]{FEFDFD}-0.1\%  & \cellcolor[HTML]{FDF5F5}-0.4\%  & \cellcolor[HTML]{FEFFFF}0.0\%    & \cellcolor[HTML]{FDF8F8}-0.2\%    & \cellcolor[HTML]{FDF6F6}-0.3\%      & \cellcolor[HTML]{FEFBFA}-0.2\%        \\
                       & W8A8    & \cellcolor[HTML]{DEF2E8}1.3\%   & \cellcolor[HTML]{FCEFEE}-0.6\%  & \cellcolor[HTML]{F9FDFB}0.3\%   & \cellcolor[HTML]{FDF6F5}-0.3\%  & \cellcolor[HTML]{FEFEFE}0.0\%    & \cellcolor[HTML]{FCFEFD}0.1\%     & \cellcolor[HTML]{F7FCFA}0.3\%       & \cellcolor[HTML]{FEFAFA}-0.2\%        \\
\multirow{-3}{*}{35B}  & W4-g     & \cellcolor[HTML]{EC9E97}-3.7\%  & \cellcolor[HTML]{F6D1CD}-1.8\%  & \cellcolor[HTML]{F6D1CE}-1.7\%  & \cellcolor[HTML]{EB9B94}-3.8\%  & \cellcolor[HTML]{EB968F}-4.0\%   & \cellcolor[HTML]{F0B0AB}-3.0\%    & \cellcolor[HTML]{F3C0BB}-2.4\%      & \cellcolor[HTML]{EB9891}-3.9\%        \\ \bottomrule
\end{tabular}%
}
\caption{\textbf{Per-language relative performance (\%$\Delta$) vs. FP16, averaged over MGSM, mMMLU, FLORES, and Language Confusion tasks.} \textbf{Ltn/IE} are Latin-script/Indo-European: de, es, fr. $\neg$ are the rest: ja, zh.}
\label{tab:avgs-with-mgsm}
\end{table}

\paragraph{Aya 23 Models} Table \ref{tab:aya-overall} shows the aggregated results for Aya 23 models on the extended Aya evaluations at \textbf{W8}, and \textbf{W4} quantization. 
We find a similar trend with Command models where \textbf{W4} often leads to a larger drop compared to \textbf{W8}, consistent across tasks and languages. \textbf{W8}, however, does not substantially drop performance in any task.

\subsection{By Task} \label{sec:results-task}
\textit{Are tasks differently affected by quantization?} 

\vspace{3pt}
Results here reference Tables \ref{tab:main-overall} and \ref{tab:aya-overall}, with full raw and relative results in Appendix \ref{sec:app-full-auto}. 
Mathematical reasoning (MGSM) is strikingly affected by quantization.  Relative performance of the 35B \textbf{W4-g} model is a dismal $-13.1\%$, with as poor as $-17.3\%$ in Chinese. MGSM and Belebele are most greatly degraded for Aya 23 models with \textbf{W4} quantization, dropping $7.5\%$ and $8.5\%$ on the 8B. mMMLU is the next most greatly degraded task. On FLORES, the 103B model is more sensitive to quantization in the L2$\rightarrow$En direction than L2$\rightarrow$En, though we see the opposite for the smaller 35B and Aya 23 models at \textbf{W4}. Quantization does not noticeably impact unseen discriminative tasks (XWinograd, XCOPA, XStoryCloze: Table \ref{tab:app-aya-unseen}).

There are some fleeting performance boosts: +1.8--2.1\% on MGSM and mild improvements on FLORES with \textbf{W8} on Aya models, and a similar translation boost of the 35B Command model at \textbf{W8A8}. Quantization generally has no effect or causes mild improvement on the monolingual language confusion task, and cross-lingual language confusion performance is boosted with greater quantization in some cases.

\subsection{By Language}
\label{sec:specific-lan}

\textit{Are languages differently affected by quantization?} 

\vspace{3pt}
Table \ref{tab:avgs-no-mgsm} averages performance over mMMLU, FLORES, and Language Confusion tasks. Table \ref{tab:avgs-with-mgsm} further includes MGSM for supported languages. Metrics are on different scales, so we average relative change (\%$\Delta$) rather than raw scores.\footnote{Ex. to arrive at $-1.3\%$ for 103B \textbf{W8A8} in Arabic, we average relative performance for mMMLU, FLORES En$\leftrightarrow$L2, and Language Confusion tasks: $\text{Avg}(\{-2.2\%,	-1.0\%,	-1.3\%,	0.0\%,	-1.8\%\}) = -1.3\%$.} We separate into languages written in the Latin/Roman script (also the subset of Indo-European languages; \textbf{Ltn/IE}) versus the rest (\textbf{$\neg$Ltn/IE}).  

\textbf{W4-g} causes considerable degradation across languages for the 35B Command model. A relationship with language is apparent: \textbf{$\neg$Ltn/IE} languages typically degrade more. Chinese, Japanese, and Korean are particularly harmed by \textbf{W4} on the 103B. The effect is seen consistently across all automatic metrics for Command, with limited exception.   Table \ref{tab:sq-g-ablation} is discussed more thoroughly in Section \ref{sec:strategies}, but also shows this discrepancy. In the Appendix, we see the same for Aya 23 models at \textbf{W4}. 

Interestingly, \textbf{W8A8} of the 35B Command model \textit{helps} on average across all languages. The magnitude is primarily due to an increase on cross-lingual language confusion. \textbf{W8} also aids Aya 23 on MGSM (Table \ref{tab:app-aya-mgsm}) for \textbf{$\neg$Ltn/IE} languages, and across languages on FLORES (Table \ref{tab:app-aya-flores}).

\begin{table*}[htb]
\centering
\resizebox{\textwidth}{!}{%
\begin{tabular}{ll|rrr|rrr|rrr|rrr}
\toprule
                       &         & \multicolumn{3}{c|}{\textbf{Avg Rel. $\%\Delta$}}                                                             & \multicolumn{3}{c|}{\textbf{mMMLU}}                                                                       & \multicolumn{3}{c|}{\textbf{MGSM}}                                                                          & \multicolumn{3}{c}{\textbf{Lang. Conf. (Mono)}}                                                              \\
                       &         & \multicolumn{1}{c}{en}         & \multicolumn{1}{c}{Ltn/IE}     & \multicolumn{1}{c|}{All}       & \multicolumn{1}{c}{en}         & \multicolumn{1}{c}{Ltn/IE}     & \multicolumn{1}{c|}{All}       & \multicolumn{1}{c}{en}         & \multicolumn{1}{c}{Ltn/IE}      & \multicolumn{1}{c|}{All}        & \multicolumn{1}{c}{en}         & \multicolumn{1}{c}{Ltn/IE}     & \multicolumn{1}{c}{All}        \\ \midrule
                       & W8      & \cellcolor[HTML]{FDFFFE}0.1\%  & \cellcolor[HTML]{FDF7F7}-0.3\% & \cellcolor[HTML]{FDF6F5}-0.3\% & \cellcolor[HTML]{FFFFFF}0.0\%  & \cellcolor[HTML]{FFFFFF}0.0\%  & \cellcolor[HTML]{FFFFFF}0.0\%  & \cellcolor[HTML]{F8FDFA}0.3\%  & \cellcolor[HTML]{FBEBEA}-0.7\%  & \cellcolor[HTML]{FAE4E3}-1.0\%  & \cellcolor[HTML]{FFFFFF}0.0\%  & \cellcolor[HTML]{FEFBFB}-0.1\% & \cellcolor[HTML]{FFFFFF}0.0\%  \\ \hhline{~*{13}{-}}
                       & W8A8-sq & \cellcolor[HTML]{F9DFDD}-1.2\% & \cellcolor[HTML]{FBEFEE}-0.6\% & \cellcolor[HTML]{FBECEA}-0.7\% & \cellcolor[HTML]{FEFBFB}-0.1\% & \cellcolor[HTML]{FDF5F5}-0.4\% & \cellcolor[HTML]{FCF0EF}-0.5\% & \cellcolor[HTML]{EDA59E}-3.4\% & \cellcolor[HTML]{F8DBD9}-1.3\%  & \cellcolor[HTML]{F6D4D1}-1.6\%  & \cellcolor[HTML]{FFFFFF}0.0\%  & \cellcolor[HTML]{FEFCFB}-0.1\% & \cellcolor[HTML]{FFFFFF}0.0\%  \\
                       & W8A8    & \cellcolor[HTML]{FDF8F7}-0.3\% & \cellcolor[HTML]{FBECEA}-0.7\% & \cellcolor[HTML]{FAE9E7}-0.8\% & \cellcolor[HTML]{FBECEB}-0.7\% & \cellcolor[HTML]{F8DBD9}-1.3\% & \cellcolor[HTML]{F6D3D0}-1.7\% & \cellcolor[HTML]{FFFFFF}0.0\%  & \cellcolor[HTML]{FAE8E6}-0.9\%  & \cellcolor[HTML]{F9E4E2}-1.0\%  & \cellcolor[HTML]{FEFCFC}-0.1\% & \cellcolor[HTML]{FEFFFF}0.0\%  & \cellcolor[HTML]{FBFEFC}0.2\%  \\ \hhline{~*{13}{-}}
                       & W4-g    & \cellcolor[HTML]{F4C9C6}-2.0\% & \cellcolor[HTML]{FAE6E4}-0.9\% & \cellcolor[HTML]{F7D7D5}-1.5\% & \cellcolor[HTML]{F6D2CE}-1.7\% & \cellcolor[HTML]{F9E2E0}-1.1\% & \cellcolor[HTML]{F7D8D6}-1.5\% & \cellcolor[HTML]{E98C84}-4.4\% & \cellcolor[HTML]{F6CFCC}-1.8\%  & \cellcolor[HTML]{EFAFA9}-3.0\%  & \cellcolor[HTML]{FFFFFF}0.0\%  & \cellcolor[HTML]{FDFEFE}0.1\%  & \cellcolor[HTML]{FFFFFF}0.0\%  \\ 
\multirow{-5}{*}{103B} & W4      & \cellcolor[HTML]{EC9D96}-3.7\% & \cellcolor[HTML]{EB9891}-3.9\% & \cellcolor[HTML]{E98D85}-4.3\% & \cellcolor[HTML]{EEA8A2}-3.3\% & \cellcolor[HTML]{EB9992}-3.9\% & \cellcolor[HTML]{E98C85}-4.4\% & \cellcolor[HTML]{E67C73}-7.9\% & \cellcolor[HTML]{E67C73}-8.0\%  & \cellcolor[HTML]{E67C73}-8.8\%  & \cellcolor[HTML]{FFFFFF}0.0\%  & \cellcolor[HTML]{FCFEFD}0.1\%  & \cellcolor[HTML]{FCFEFD}0.1\%  \\ \midrule\midrule
                       & W8      & \cellcolor[HTML]{FEFCFC}-0.1\% & \cellcolor[HTML]{FEF9F9}-0.2\% & \cellcolor[HTML]{FDF8F8}-0.2\% & \cellcolor[HTML]{FEFDFD}-0.1\% & \cellcolor[HTML]{FEFDFD}-0.1\% & \cellcolor[HTML]{FEFDFD}-0.1\% & \cellcolor[HTML]{FDF7F7}-0.3\% & \cellcolor[HTML]{FCEFEE}-0.6\%  & \cellcolor[HTML]{FBEAE9}-0.8\%  & \cellcolor[HTML]{FDFEFE}0.1\%  & \cellcolor[HTML]{FDFFFE}0.1\%  & \cellcolor[HTML]{FDFFFE}0.1\%  \\
                       & W8A8    & \cellcolor[HTML]{F9FDFB}0.2\%  & \cellcolor[HTML]{F6D4D1}-1.6\% & \cellcolor[HTML]{F5CFCC}-1.8\% & \cellcolor[HTML]{FFFFFF}0.0\%  & \cellcolor[HTML]{FEFEFE}0.0\%  & \cellcolor[HTML]{FEFAF9}-0.2\% & \cellcolor[HTML]{FFFFFF}0.0\%  & \cellcolor[HTML]{E67D75}-4.9\%  & \cellcolor[HTML]{E67C73}-5.6\%  & \cellcolor[HTML]{EDF8F3}0.7\%  & \cellcolor[HTML]{FFFFFF}0.0\%  & \cellcolor[HTML]{F7FCF9}0.3\%  \\
\multirow{-3}{*}{35B}  & W4-g    & \cellcolor[HTML]{F9DFDD}-1.2\% & \cellcolor[HTML]{E78279}-4.8\% & \cellcolor[HTML]{E67C73}-5.2\% & \cellcolor[HTML]{F6D0CD}-1.8\% & \cellcolor[HTML]{F7D7D5}-1.5\% & \cellcolor[HTML]{F4CAC6}-2.0\% & \cellcolor[HTML]{F4C5C1}-2.2\% & \cellcolor[HTML]{E67C73}-12.2\% & \cellcolor[HTML]{E67C73}-13.1\% & \cellcolor[HTML]{F5FBF8}0.4\%  & \cellcolor[HTML]{FBEEEC}-0.6\% & \cellcolor[HTML]{FCF3F2}-0.4\% \\ \bottomrule
\end{tabular}%
}
\caption{\textbf{Relative performance of quantized Command models in English vs. other languages.} All non-English languages (All), non-English Latin-script/Indo-European languages (Ltn/IE).}
\label{tab:results-vs-en-command}
\end{table*}

\pagebreak
\textit{How does training data size affect performance?} 

\begin{figure}[htb]
    \centering
    \includegraphics[height=5.0cm, width=1\linewidth]{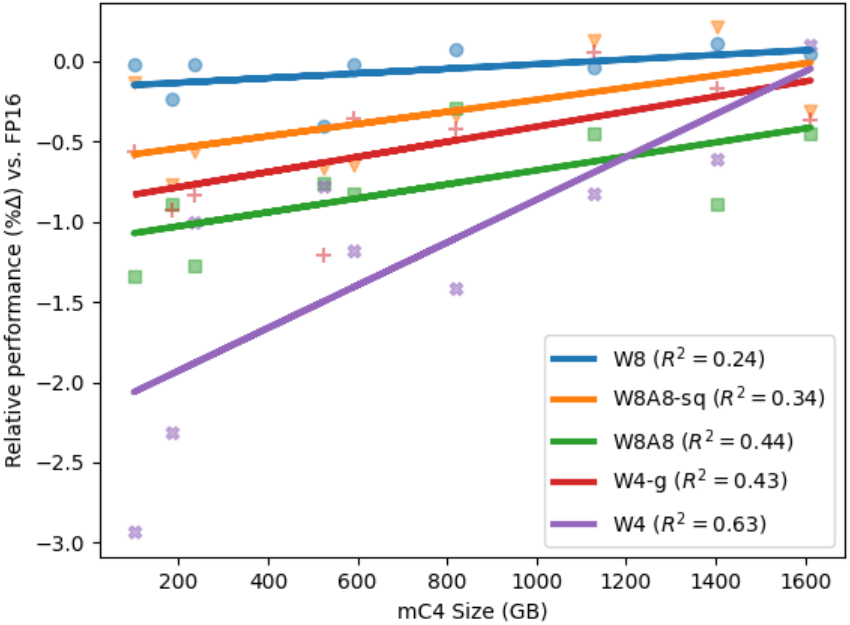}
    \caption{\textbf{Data size in mC4 \citep{xue-etal-2021-mt5} vs. avg. perf. under quantization.} Table \ref{tab:avgs-no-mgsm}, Command 103B.}
    \label{fig:perf-v-mc4size}
\end{figure}

\vspace{3pt}
Training mixtures for Command and Aya 23 models are not released, so a definitive relationship between data set size and downstream performance cannot be determined.  Instead, we might assume the training data follows the distribution from well-known large multilingual corpora. 
In Figure \ref{fig:perf-v-mc4size}, we correlate per-language relative performance vs. FP16 from Table \ref{tab:avgs-no-mgsm} with amount of data in mC4 \citep{xue-etal-2021-mt5}.\footnote{\url{https://github.com/allenai/allennlp/discussions/5265}.  
Correlation with size in tokens from \citet{xue-etal-2021-mt5}'s Table 6 shows similar 
 $R^2$.  
Data size by lang. (GB): [ar: 57, de: 347, es: 433, fr: 318, it: 162, ja: 164, ko: 26, pt: 146, zh: 39].
English excluded as it cannot be averaged with FLORES \& cross-lingual Language Confusion.} The correlation between downstream performance and data set size is stronger as quantization becomes more extreme: from $R^2 = 0.24$ for \textbf{W8} to $R^2 = 0.63$ with \textbf{W4}.

\textit{How does performance compare to English?} 
\vspace{3pt}

Table \ref{tab:results-vs-en-command} shows relative performance of quantized Command 103B and 35B models in English vs. other languages for tasks which could be evaluated in English.\footnote{FLORES / cross-lingual Language Confusion cannot be.} Under most settings, English degrades less than the average of all others.  The largest gap is on MGSM for the 35B, where the model is very sensitive to \textbf{W8A8} and \textbf{W4-g} quantization outside of English.  Results for the Aya 23 models are in Table \ref{tab:app-results-vs-en-aya}, where performance is worse on average for non-English languages at \textbf{W4}, while being less consistent at \textbf{W8}.

\subsection{By Model Size} \label{sec:results-model-size}

\textit{How do model size and quantization level interact?}

\vspace{3pt}
Across evaluations at the most extreme quantization (\textbf{W4}/\textbf{W4-g}), smaller models are more sensitive: \textbf{W4-g} variants of 103B and 35B Command record $-0.9\%$ and $-2.8\%$ performance relative to FP16 on average, with a stark difference of $-2.9\%$ vs. $-13.1\%$ on MGSM. Aya 23 35B/8B record $-2.9\%$ vs. $-3.7\%$ on average, with their largest gap occurring in Belebele ($-5.9\%$ vs. $-8.5\%$). (Refer back to Tables \ref{tab:main-overall} and \ref{tab:aya-overall}.)

\subsection{By Quantization Strategy}
\label{sec:strategies}

\textit{How do techniques like SmoothQuant and group-wise scaling affect downstream performance?}
\vspace{3pt}

\begin{table*}[h]
\centering
\resizebox{\textwidth}{!}{%
\begin{tabular}{l|rr|rr|rr|rr|rr|rr|rr}
\toprule
 &  &   &    &    &    &    &
  \multicolumn{4}{c|}{\textbf{FLORES}} &
  \multicolumn{4}{c}{\textbf{Language Confusion}} \\ 
 &
  \multicolumn{2}{c|}{\textbf{Avg. Rel. \%}} &
  \multicolumn{2}{c|}{\textbf{mMMLU}} &
  \multicolumn{2}{c|}{\textbf{MGSM}} &
  \multicolumn{2}{c|}{\textbf{En $\rightarrow$ L2}} &
  \multicolumn{2}{c|}{\textbf{L2 $\rightarrow$ En}} &
  \multicolumn{2}{c|}{\textbf{Monolingual}} &
  \multicolumn{2}{c}{\textbf{Cross-lingual}} \\ 
 &
  \multicolumn{1}{l}{Ltn/IE} &
  \multicolumn{1}{c|}{$\neg$} &
  \multicolumn{1}{l}{Ltn/IE} &
  \multicolumn{1}{c|}{$\neg$} &
  \multicolumn{1}{l}{Ltn/IE} &
  \multicolumn{1}{c|}{$\neg$} &
  \multicolumn{1}{l}{Ltn/IE} &
  \multicolumn{1}{c|}{$\neg$} &
  \multicolumn{1}{l}{Ltn/IE} &
  \multicolumn{1}{c|}{$\neg$} &
  \multicolumn{1}{l}{Ltn/IE} &
  \multicolumn{1}{c|}{$\neg$} &
  \multicolumn{1}{l}{Ltn/IE} &
  \multicolumn{1}{c}{$\neg$} \\ \midrule
W8A8    & \cellcolor[HTML]{FBECEB}-0.7\% & \cellcolor[HTML]{F9E4E2}-1.0\% & \cellcolor[HTML]{F8DBD9}-1.3\% & \cellcolor[HTML]{F4C8C4}-2.1\% & \cellcolor[HTML]{FAE8E6}-0.9\% & \cellcolor[HTML]{F8DDDB}-1.3\%  & \cellcolor[HTML]{FEFBFB}-0.1\% & \cellcolor[HTML]{FCFEFD}0.1\%  & \cellcolor[HTML]{F9E4E2}-1.0\% & \cellcolor[HTML]{F7D5D2}-1.6\% & \cellcolor[HTML]{FEFFFF}0.0\%  & \cellcolor[HTML]{F6FCF9}0.4\%  & \cellcolor[HTML]{FAE6E4}-0.9\% & \cellcolor[HTML]{F7D6D3}-1.6\% \\
W8A8-sq & \cellcolor[HTML]{FCF3F2}-0.4\% & \cellcolor[HTML]{FBECEB}-0.7\% & \cellcolor[HTML]{FDF5F5}-0.4\% & \cellcolor[HTML]{FBEAE9}-0.8\% & \cellcolor[HTML]{F8DBD9}-1.3\% & \cellcolor[HTML]{F5CDCA}-1.9\%  & \cellcolor[HTML]{FAFDFB}0.2\%  & \cellcolor[HTML]{FEFEFD}0.0\%  & \cellcolor[HTML]{F9E2E0}-1.1\% & \cellcolor[HTML]{F6D4D1}-1.6\% & \cellcolor[HTML]{FEFCFB}-0.1\% & \cellcolor[HTML]{FCFEFD}0.1\%  & \cellcolor[HTML]{FEFFFE}0.1\%  & \cellcolor[HTML]{FEFFFF}0.0\%  \\ \midrule
W4      & \cellcolor[HTML]{F5CDCA}-1.9\% & \cellcolor[HTML]{EEA8A2}-3.3\% & \cellcolor[HTML]{EB9992}-3.9\% & \cellcolor[HTML]{E67D74}-4.9\% & \cellcolor[HTML]{E67C73}-8.0\% & \cellcolor[HTML]{E67C73}-10.2\% & \cellcolor[HTML]{F8DCDA}-1.3\% & \cellcolor[HTML]{F4C9C5}-2.0\% & \cellcolor[HTML]{F9E1DF}-1.1\% & \cellcolor[HTML]{F3C2BD}-2.3\% & \cellcolor[HTML]{FCFEFD}0.1\%  & \cellcolor[HTML]{FCFEFD}0.1\%  & \cellcolor[HTML]{B4E1CB}2.9\%  & \cellcolor[HTML]{FCF3F3}-0.4\% \\
W4-g     & \cellcolor[HTML]{FBEEED}-0.6\% & \cellcolor[HTML]{F8DAD8}-1.4\% & \cellcolor[HTML]{F9E2E0}-1.1\% & \cellcolor[HTML]{F5CCC9}-1.9\% & \cellcolor[HTML]{F6CFCC}-1.8\% & \cellcolor[HTML]{E67E75}-4.9\%  & \cellcolor[HTML]{FBFEFC}0.2\%  & \cellcolor[HTML]{ECF8F2}0.7\%  & \cellcolor[HTML]{FDF6F5}-0.3\% & \cellcolor[HTML]{FAE9E7}-0.8\% & \cellcolor[HTML]{FDFEFE}0.1\%  & \cellcolor[HTML]{FEFCFC}-0.1\% & \cellcolor[HTML]{FAE7E5}-0.9\% & \cellcolor[HTML]{F8DCD9}-1.3\% \\ \bottomrule
\end{tabular}%
}
\caption{\textbf{Effect of mitigation strategies on W8A8 and W4 quantization on the 103B model.} Percentage points off FP16 baseline for W8A8-sq vs. naive W8A8 and W4-g vs. W4, broken down by Latin-script/Indo-European languages (Ltn/IE) versus others ($\neg$). Avg. Rel. \% reports averaged performance all datasets.}
\label{tab:sq-g-ablation}
\end{table*}

\begin{table*}[htb]
\centering
\resizebox{\textwidth}{!}{%
\begin{tabular}{@{}ll|rr|rr|rr|rr||rr|rr|rr}
\toprule
 &
   &
  \multicolumn{2}{c}{\textbf{fr}} &
  \multicolumn{2}{c}{\textbf{es}} &
  \multicolumn{2}{c}{\textbf{ja}} &
  \multicolumn{2}{c||}{\textbf{ko}} &
  \multicolumn{2}{c}{\textbf{Avg}} &
  \multicolumn{2}{c}{\textbf{Ltn/IE}} &
  \multicolumn{2}{c}{\textbf{$\neg$}} \\
\textbf{} &
  \multicolumn{1}{l|}{\textbf{}} &
  \multicolumn{1}{c}{\textbf{LLM}} &
  \multicolumn{1}{c|}{\textbf{RM}} &
  \multicolumn{1}{c}{\textbf{LLM}} &
  \multicolumn{1}{c|}{\textbf{RM}} &
  \multicolumn{1}{c}{\textbf{LLM}} &
  \multicolumn{1}{c|}{\textbf{RM}} &
  \multicolumn{1}{c}{\textbf{LLM}} &
  \multicolumn{1}{c||}{\textbf{RM}} &
  \multicolumn{1}{c}{\textbf{LLM}} &
  \multicolumn{1}{c|}{\textbf{RM}} &
  \multicolumn{1}{c}{\textbf{LLM}} &
  \multicolumn{1}{c|}{\textbf{RM}} &
  \multicolumn{1}{c}{\textbf{LLM}} &
  \multicolumn{1}{c}{\textbf{RM}} \\ \midrule
&
  W8 &
  \cellcolor[HTML]{F3FAF7}1.0\% &
  \cellcolor[HTML]{FEFAFA}-0.7\% &
  \cellcolor[HTML]{F2BCB7}-10.2\% &
  \cellcolor[HTML]{A1D9BE}7.5\% &
  \cellcolor[HTML]{F8DBD9}-5.4\% &
  \cellcolor[HTML]{BBE4D0}5.4\% &
  \cellcolor[HTML]{A2D9BE}7.5\% &
  \cellcolor[HTML]{F7D9D6}-5.8\% &
  \cellcolor[HTML]{FCF3F2}-1.8\% &
  \cellcolor[HTML]{EBF7F1}1.6\% &
  \cellcolor[HTML]{F9E1DF}-4.6\% &
  \cellcolor[HTML]{D5EEE2}3.4\% &
  \cellcolor[HTML]{F3FAF7}1.0\% &
  \cellcolor[HTML]{FEFDFD}-0.2\% \\
 &
  W8A8-sq &
  \cellcolor[HTML]{E8867E}-18.4\% &
  \cellcolor[HTML]{F8DDDB}-5.1\% &
  \cellcolor[HTML]{FAE6E4}-3.7\% &
  \cellcolor[HTML]{CCEBDC}4.1\% &
  \cellcolor[HTML]{E6F5EE}2.0\% &
  \cellcolor[HTML]{C4E7D6}4.7\% &
  \cellcolor[HTML]{D0ECDF}3.7\% &
  \cellcolor[HTML]{F8DDDB}-5.1\% &
  \cellcolor[HTML]{F9E4E2}-4.1\% &
  \cellcolor[HTML]{FEFCFC}-0.3\% &
  \cellcolor[HTML]{F1B6B1}-11.0\% &
  \cellcolor[HTML]{FEFBFB}-0.5\% &
  \cellcolor[HTML]{DBF1E6}2.9\% &
  \cellcolor[HTML]{FEFDFD}-0.2\% \\
 &
  W4-g &
  \cellcolor[HTML]{F1BAB5}-10.5\% &
  \cellcolor[HTML]{E98F88}-17.0\% &
  \cellcolor[HTML]{EA928A}-16.6\% &
  \cellcolor[HTML]{E6F5EE}2.0\% &
  \cellcolor[HTML]{EB9A93}-15.3\% &
  \cellcolor[HTML]{FFFFFF}0.0\% &
  \cellcolor[HTML]{F7D9D6}-5.8\% &
  \cellcolor[HTML]{EB9891}-15.6\% &
  \cellcolor[HTML]{EFB0AA}-12.1\% &
  \cellcolor[HTML]{F5CCC9}-7.7\% &
  \cellcolor[HTML]{EEA6A0}-13.6\% &
  \cellcolor[HTML]{F5CECA}-7.5\% &
  \cellcolor[HTML]{F1B9B5}-10.5\% &
  \cellcolor[HTML]{F5CBC8}-7.8\% \\
\multirow{-4}{*}{Internal} &
  W4 &
  \cellcolor[HTML]{E67C73}-30.2\% &
  \cellcolor[HTML]{E67C73}-20.4\% &
  \cellcolor[HTML]{E67C73}-33.0\% &
  \cellcolor[HTML]{E98F88}-17.0\% &
  \cellcolor[HTML]{E67C73}-21.7\% &
  \cellcolor[HTML]{E67C73}-20.0\% &
  \cellcolor[HTML]{E7847C}-18.6\% &
  \cellcolor[HTML]{E67C73}-27.6\% &
  \cellcolor[HTML]{E67C73}-25.9\% &
  \cellcolor[HTML]{E67C73}-21.2\% &
  \cellcolor[HTML]{E67C73}-31.6\% &
  \cellcolor[HTML]{E7847C}-18.7\% &
  \cellcolor[HTML]{E67C73}-20.2\% &
  \cellcolor[HTML]{E67C73}-23.8\% \\ \midrule
&
  W8 &
  \cellcolor[HTML]{FDF6F5}-1.3\% &
  \cellcolor[HTML]{E6F5EE}2.0\% &
  \cellcolor[HTML]{A3DABF}7.3\% &
  \cellcolor[HTML]{FAE4E2}-4.0\% &
  \cellcolor[HTML]{F7D7D5}-6.0\% &
  \cellcolor[HTML]{F8DCD9}-5.3\% &
  \cellcolor[HTML]{DEF2E8}2.7\% &
  \cellcolor[HTML]{E6F5EE}2.0\% &
  \cellcolor[HTML]{F7FCFA}0.7\% &
  \cellcolor[HTML]{FDF6F5}-1.3\% &
  \cellcolor[HTML]{DAF0E5}3.0\% &
  \cellcolor[HTML]{FDF8F8}-1.0\% &
  \cellcolor[HTML]{FCF4F3}-1.7\% &
  \cellcolor[HTML]{FCF4F3}-1.7\% \\
 &
  W8A8-sq &
  \cellcolor[HTML]{EB9A93}-15.3\% &
  \cellcolor[HTML]{F4C6C2}-8.7\% &
  \cellcolor[HTML]{92D3B4}8.7\% &
  \cellcolor[HTML]{F5CAC7}-8.0\% &
  \cellcolor[HTML]{FDF6F5}-1.3\% &
  \cellcolor[HTML]{EFF9F4}1.3\% &
  \cellcolor[HTML]{F5CAC7}-8.0\% &
  \cellcolor[HTML]{F9E0DE}-4.7\% &
  \cellcolor[HTML]{F9E4E2}-4.0\% &
  \cellcolor[HTML]{F8DEDC}-5.0\% &
  \cellcolor[HTML]{FAE9E7}-3.3\% &
  \cellcolor[HTML]{F4C8C4}-8.3\% &
  \cellcolor[HTML]{F9E0DE}-4.7\% &
  \cellcolor[HTML]{FCF4F3}-1.7\% \\
 &
  W8A8 &
  \cellcolor[HTML]{FAE8E7}-3.4\% &
  \cellcolor[HTML]{DEF2E8}2.7\% &
  \cellcolor[HTML]{57BB8A}13.3\% &
  \cellcolor[HTML]{FAE9E7}-3.3\% &
  \cellcolor[HTML]{DEF2E8}2.7\% &
  \cellcolor[HTML]{FDF6F5}-1.3\% &
  \cellcolor[HTML]{BCE4D1}5.3\% &
  \cellcolor[HTML]{FAE9E7}-3.3\% &
  \cellcolor[HTML]{C7E9D8}4.5\% &
  \cellcolor[HTML]{FDF6F5}-1.3\% &
  \cellcolor[HTML]{C1E6D4}5.0\% &
  \cellcolor[HTML]{FEFCFC}-0.3\% &
  \cellcolor[HTML]{CDEBDC}4.0\% &
  \cellcolor[HTML]{FCEFEE}-2.3\% \\
\multirow{-4}{*}{Dolly} &
  W4-g &
  \cellcolor[HTML]{F5CECB}-7.4\% &
  \cellcolor[HTML]{FBEDEC}-2.7\% &
  \cellcolor[HTML]{FAE4E2}-4.0\% &
  \cellcolor[HTML]{C5E8D7}4.7\% &
  \cellcolor[HTML]{EB9A93}-15.3\% &
  \cellcolor[HTML]{EB9A93}-15.3\% &
  \cellcolor[HTML]{F0B4AF}-11.3\% &
  \cellcolor[HTML]{F8DCD9}-5.3\% &
  \cellcolor[HTML]{F3C0BC}-9.5\% &
  \cellcolor[HTML]{F9E0DE}-4.7\% &
  \cellcolor[HTML]{F7D9D7}-5.7\% &
  \cellcolor[HTML]{F3FAF7}1.0\% &
  \cellcolor[HTML]{EEA7A1}-13.3\% &
  \cellcolor[HTML]{F2BBB6}-10.3\% \\ \bottomrule 
\end{tabular}%
}
\caption{\textbf{Relative performance vs. FP16 of 103B quantized models according to \textit{LLM/RM-as-a-Judge}} over \textit{Internal} and \textit{Aya Dolly} subsampled test sets. Raw win-rates in Table \ref{tab:app-llm-as-a-judge-raw}.}
\label{tab:llm-as-a-judge-rel}
\end{table*}

Table \ref{tab:sq-g-ablation} shows the effect of using SmoothQuant and Group-Wise scaling strategies.
We evaluate variants of the 103B Command model with SmoothQuant (\textbf{W8A8-sq}), and a more naive \textbf{W4} variant using per-column quantization instead of group-wise scaling.  We compare \textbf{W8A8-sq} to \textbf{W8A8}, and \textbf{W4-g} to \textbf{W4}. 

On average and across mMMLU, MGSM, and FLORES, Group-Wise scaling greatly improves over column-wise \textbf{W4}, recovering over 6 percentage points lost on MGSM for \textbf{Ltn/IE} languages. SmoothQuant has a similar effect on average and for mMMLU, though to a lesser degree. That said, SmoothQuant harms MGSM scores slightly, and Group-Wise scaling degrades cross-lingual language confusion.  We again observe that \textbf{$\neg\text{Ltn/IE}$} languages suffer more in nearly all cases.

On cross-lingual language confusion, strategies aimed to retain performance have different effects: SmoothQuant recovers all lost from naive \textbf{W8A8}, but Group-Wise scaling is actively damaging. In contrast, \textbf{W4} benefits \textbf{Ltn/IE} and Arabic on cross-lingual language confusion, but worsens the rest.\footnote{Full results in are Table \ref{tab:app-langconf-pct}.} 

Thus, while the quantization strategies tend to aid performance overall, there may be adverse effects on specific tasks. 
More research is needed to understand this, but it is intriguing to consider the effect that lower-precision might have on the ability to produce output in a desired language, and maintain that language once decoding begins.

\subsection{LLM/RM-as-a-Judge} \label{sec:results-llm-as-a-judge}

Table \ref{tab:llm-as-a-judge-rel} shows relative performance of quantized variants of the 103B Command model evaluated with LLM- and RM-as-a-Judge.\footnote{Calculation: $\frac{\text{Quantized Win Rate} - 50}{50}$, as 50 is the expected win-rate of two FP16 models compared.}  In nearly all cases, the LLM and RM agree that \textbf{W4} and \textbf{W4-g} severely harm performance on our challenging \textit{Internal} test set.  Performance is also severely degraded for \textbf{$\neg$Ltn/IE} languages on \textit{Dolly} with \textbf{W4-g}, and French with \textbf{W8A8-sq}. On average across languages, the LLM and RM agree on the ranking of model quality over \textit{Internal}.  Results on the easier \textit{Dolly} test set are less clear-cut: The LLM reports greater degradation for \emph{Internal} than \textit{Dolly} overall, but the RM disagrees for \textbf{W8} and \textbf{W8A8-sq}.  Perhaps \textit{Dolly} prompts are easy enough that models output similar responses, creating more noise in the judgments; future work could examine this hypothesis.  Furthermore, on multiple instances, the LLM and RM disagree on whether performance \textit{improves} or \textit{worsens}, given the same setting. Comparisons between the two differing methods of automated evaluation are worthy of further study.

\subsection{Human Evaluation} \label{sec:results-human-evaluation}

\begin{table}[htb]
\resizebox{0.47\textwidth}{!}{%
\begin{tabular}{@{}ll|rrrr||r|rrr@{}}
\toprule
 &  & \multicolumn{1}{l}{} & \multicolumn{1}{l}{} & \multicolumn{1}{l}{} & \multicolumn{1}{l|}{} & \multicolumn{1}{l|}{} & \multicolumn{3}{c}{non-English Stats} \\ \midrule
 & \multicolumn{1}{l|}{} & \multicolumn{1}{c}{fr} & \multicolumn{1}{c}{es} & \multicolumn{1}{c}{ja} & \multicolumn{1}{c|}{ko} & \multicolumn{1}{c|}{en} & \multicolumn{1}{c}{avg} & \multicolumn{1}{c}{Ltn/IE} & \multicolumn{1}{c}{$\neg$} \\ \midrule
 & \multicolumn{1}{l|}{W8} & \cellcolor[HTML]{F5CECB}-7.4\% & \cellcolor[HTML]{F7FCFA}0.6\% & \cellcolor[HTML]{98D6B7}7.4\% & \multicolumn{1}{r|}{\cellcolor[HTML]{F0B0AB}-12.0\%} & \multicolumn{1}{r|}{\cellcolor[HTML]{FAE4E2}-4.0\%} & \cellcolor[HTML]{FBECEB}-2.8\% & \cellcolor[HTML]{FAE8E7}-3.4\% & \cellcolor[HTML]{FCEFEE}-2.3\% \\
 & \multicolumn{1}{l|}{W8A8-sq} & \cellcolor[HTML]{F3C1BD}-9.4\% & \cellcolor[HTML]{F5CECB}-7.4\% & \cellcolor[HTML]{FCF1F1}-2.0\% & \multicolumn{1}{r|}{\cellcolor[HTML]{C7E9D8}4.0\%} & \multicolumn{1}{r|}{\cellcolor[HTML]{A3DABF}6.6\%} & \cellcolor[HTML]{FAE6E5}-3.7\% & \cellcolor[HTML]{F4C7C4}-8.4\% & \cellcolor[HTML]{F2FAF6}1.0\% \\
\multirow{-3}{*}{Internal} & \multicolumn{1}{l|}{W4-g} & \cellcolor[HTML]{EA928A}-16.6\% & \cellcolor[HTML]{F9E0DE}-4.6\% & \cellcolor[HTML]{EB968F}-16.0\% & \multicolumn{1}{r|}{\cellcolor[HTML]{F9E0DE}-4.6\%} & \multicolumn{1}{r|}{\cellcolor[HTML]{F5CECB}-7.4\%} & \cellcolor[HTML]{F1BAB5}-10.5\% & \cellcolor[HTML]{F1B9B4}-10.6\% & \cellcolor[HTML]{F2BBB6}-10.3\% \\ \midrule
 & \multicolumn{1}{l|}{W8} & \cellcolor[HTML]{F7FCFA}0.6\% & \cellcolor[HTML]{F8DBD9}-5.4\% & \cellcolor[HTML]{57BB8A}12.0\% & \multicolumn{1}{r|}{\cellcolor[HTML]{FFFFFF}0.0\%} & \multicolumn{1}{r|}{\cellcolor[HTML]{F7D7D5}-6.0\%} & \cellcolor[HTML]{E6F5EE}1.8\% & \cellcolor[HTML]{FBEFEE}-2.4\% & \cellcolor[HTML]{ABDDC5}6.0\% \\
 & \multicolumn{1}{l|}{W8A8-sq} & \cellcolor[HTML]{F5CECB}-7.4\% & \cellcolor[HTML]{F4C6C2}-8.6\% & \cellcolor[HTML]{FFFFFF}0.0\% & \multicolumn{1}{r|}{\cellcolor[HTML]{FAE8E7}-3.4\%} & \multicolumn{1}{r|}{\cellcolor[HTML]{E3F4EC}2.0\%} & \cellcolor[HTML]{F8DFDD}-4.8\% & \cellcolor[HTML]{F5CAC7}-8.0\% & \cellcolor[HTML]{FCF3F3}-1.7\% \\
\multirow{-3}{*}{Dolly} & \multicolumn{1}{l|}{W4-g} & \cellcolor[HTML]{F3C1BD}-9.4\% & \cellcolor[HTML]{FDF5F5}-1.4\% & \cellcolor[HTML]{DBF1E6}2.6\% & \multicolumn{1}{r|}{\cellcolor[HTML]{F5CAC7}-8.0\%} & \multicolumn{1}{r|}{\cellcolor[HTML]{F2BDB9}-10.0\%} & \cellcolor[HTML]{F9E4E2}-4.1\% & \cellcolor[HTML]{F8DBD9}-5.4\% & \cellcolor[HTML]{FBEDEC}-2.7\% \\ \bottomrule
\end{tabular}%
}
\caption{\textbf{Relative performance vs.~FP16 of 103B quantized models according to \textit{human evaluators}} over \textit{Internal} and \textit{Aya Dolly} subsampled test sets.} 
\label{tab:human-eval-rel}
\end{table}

Human evaluation paints a similar picture in Table~\ref{tab:human-eval-rel}, with some outliers. Average performance drops steadily across evaluated languages on the \emph{Internal} test set, which has more difficult prompts. The sharpest decline is in French, with $-16.6\%$ at \textbf{W4-g}. Curiously, there is an initial $7.4\%$ boost for Japanese with \textbf{W8}, but it falls to $-16.0\%$ with more extreme quantization. Interestingly, human annotators generally prefer outputs of quantized models on \textit{Dolly} prompts in Japanese, too, but disprefer those in other languages. We see more pronounced degradation on \emph{Internal} overall, with an average relative drop of 5.7\% versus 2.4\% for Dolly. 

\section{Related Work}

\paragraph{Impact of Compression on Multilingual Tasks}
There is a scarcity of research examining the impact of compression and quantization on multilingual tasks. \citet{paglieri2024outliers} examine multilingual calibration sets, but their evaluation is English-only. \citet{ramesh-etal-2023-comparative} study compression vis-a-vis multilingual fairness, showing that performance differs across languages and dimensions. \citet{kharazmi2023distill} show that recovering compression-caused performance loss of LSTMs is harder multilingually than monolingually. In machine translation,  distillation has varied effects by language related to priors such as amount of synthetic data used and confidence of the teacher models, while quantization exhibits more consistent trends across languages \citep{diddee-etal-2022-brittle}. To our knowledge, ours is the first to study quantized LLMs for open-ended multilingual generation.

Multilingual data is an example of long tail data. Prior work shows that compression techniques like quantization and sparsity amplify disparate treatment of long-tail rare features \citep[e.g.][]{Hooker2019WhatDC,ahia2021lowresource,ogueji-etal-2022-intriguing,hooker2020characterising}. 
Similar to our observation of occasional performance gain, \citet{ogueji-etal-2022-intriguing} show that
sparsity-based compression sometimes makes a model better suited to the downstream task. \citet{ahia2021lowresource} find that sparsity preserves machine translation performance on frequent sentences, but disparately impacts infrequent sentences. \citet{badshah2024quantifying} also report some performance gain at lower precision.

\paragraph{Quantization of LLMs} 
Recent work to improve quantized LLMs solely focuses on English models and data for tuning and evaluation \citep[e.g.][]{ahmadian-intruiging, dettmers2022llm, xiao2023smoothquant, bondarenko2024quantizable, gong2024makes}. \citet{dettmers2023case} perform a fine-grained sweep across bit-widths (3-8 bit), data types and quantization methods, and recommended 4-bit as the optimal size-performance trade-off, but do not evaluate multilingually.  \citet{huang2024empiricalstudyllama3quantization} extensively analyze quantized LLaMA3 models in English. 
\citet{badshah2024quantifying} examine the effect of 4-bit NormalFloat \citep{dettmers2023qlora} and 8-bit LLM.int8() \cite{dettmers2022llm} on across model sizes and a variety of English tasks, finding that larger models are more resilient to quantization and performs better than smaller models at higher precision. 
Even the most recent \citep[e.g.][]{li2024evaluating, liu-etal-2024-emergent-abilities} omit multilinguality without acknowledging the limitation. 

\paragraph{Model design choices} We consider how design choices like quantization impact performance for users of different languages. A wider body of work examines how design choices impact performance on underrepresented features or subgroups. \citet{zhuang2021randomness} and \citet{nelaturu2023fairness} find that hardware choice incurs disparate impact on underrepresented features.  \citet{wang2022robust} show that distillation imposes similar trade-offs, and that harm to the long-tail can be mitigated by modifying the student-teacher objective. 
\citet{ko2023fairensemble} show that ensembling disproportionately favors underrepresented attributes.  Differential privacy techniques like gradient clipping and noise injection also disproportionately impact underrepresented features \citep{bagdasaryan2019differential}.

\section{Conclusion \& Future Work}
We examine widely adopted quantization techniques for model compression and ask, \textit{How does quantization impact different languages?}  We perform an extensive study in state-of-the-art multilingual LLMs---from 8 billion to 103 billion parameters---in 20+ languages using automatic metrics, LLM/RM-as-a-Judge, and human evaluation.  We find that: (1) Damage from quantization is much worse than appears from automatic metrics: even when not observed automatically, human evaluators notice it. (2) Quantization affects languages to varying degrees, with non-Latin script languages more severely affected on automatic benchmarks. (3) Challenging tasks degrade fast and severely (e.g. mathematical reasoning and responses to realistic challenging prompts).  On a bright note, quantization occasionally brings performance benefits. 

Our results urge attention to multilingual performance at all stages of system design and might be extended to consider, for instance, languages excluded from training and out-of-distribution tasks. By minding the impact on long-tail features, we’ll build better systems to serve the world.

\section{Limitations}

\paragraph{Generality of findings} Due to the number of methods, languages, and benchmarks we examine, we focus our evaluation on models from two families (Command R/R+ and Aya 23). As we observe similar trends across these models, our findings are likely to generalize to other LLMs. Nevertheless, models that have been optimized differently or trained with a focus on specific tasks such as code or mathematical reasoning may behave differently.

\paragraph{Under-represented languages} For our study, we focused on languages that were supported by the models we evaluated. Performance deterioration is likely even larger for languages that are not in the pre-training data, or are severely under-represented. For such languages, evaluation is also more challenging due to poor availability of benchmark data and human annotators.

\bibliography{main}

\clearpage
\newpage
\appendix
\setcounter{table}{0}
\setcounter{figure}{0}
\renewcommand{\thetable}{A\arabic{table}}
\renewcommand{\thefigure}{A\arabic{figure}}

\section{Appendix}

\subsection{Prompts for mMMLU and LLM-as-a-Judge}
\label{sec:app-llm-as-a-judge}

\begin{table}[htb]
\centering
    \resizebox{\columnwidth}{!}{

    }
\caption{\textbf{mMMLU prompt}. Following \citet{achiam2023gpt}, letter choices and ``Answer'' are kept in English.}
\label{tab:app-mmmlu-prompt}
\end{table}

\begin{table*}[!htb]
    \centering                                    
    \small
    %
    
\caption{\textbf{Example Input for LLM-as-a-Judge.} Template derived from \citet{alpaca_eval}: \url{https://github.com/tatsu-lab/alpaca_eval/blob/main/src/alpaca_eval/evaluators_configs/gpt-3.5-turbo-1106_ranking/ranking_prompt.txt}}
\label{tab:app-llm-as-a-judge-prompt-gpt4}
\end{table*}

\clearpage
\newpage
\onecolumn
\subsection{Automatic Tasks - Full Results}
\label{sec:app-full-auto}
\begin{table*}[htb]
\footnotesize
\centering
\begin{tabular}{ll|rrrrr|ra}
\toprule
 &
   &
  \multicolumn{1}{c}{\textbf{de}} &
  \multicolumn{1}{c}{\textbf{es}} &
  \multicolumn{1}{c}{\textbf{fr}} &
  \multicolumn{1}{c}{\textbf{ja}} &
  \multicolumn{1}{c}{\textbf{zh}} &
  \multicolumn{1}{|c}{\textbf{en}} &
  \multicolumn{1}{c}{\textbf{non-en avg}} \\ \midrule
\multirow{6}{*}{103B} & FP16    & 72.6 & 76.6 & 70.6 & 63.0 & 70.2 & 84.0 & 70.6 \\
                      & W8      & 72.8 & 75.9 & 69.5 & 61.3 & 70.2 & 84.2 & 69.9 \\
                      & W8A8-sq & 73.4 & 73.8 & 69.6 & 62.9 & 67.7 & 81.1 & 69.5 \\
                      & W8A8    & 74.1 & 73.9 & 69.8 & 63.4 & 68.0 & 84.0 & 69.8 \\
                      & W4-g    & 71.2 & 75.7 & 69.0 & 58.0 & 68.9 & 80.3 & 68.6 \\
                      & W4      & 64.6 & 71.3 & 66.5 & 56.1 & 63.5 & 77.4 & 64.4 \\ \midrule
\multirow{4}{*}{35B}  & FP16    & 56.6 & 57.3 & 51.8 & 38.8 & 44.4 & 58.5 & 49.8 \\
                      & W8      & 55.9 & 56.6 & 52.1 & 37.4 & 45.1 & 58.3 & 49.4 \\
                      & W8A8    & 54.2 & 53.4 & 49.9 & 35.8 & 42.0 & 58.5 & 47.1 \\
                      & W4-g    & 47.2 & 51.0 & 47.1 & 34.3 & 36.7 & 57.2 & 43.3 \\ \bottomrule
\end{tabular}%
\caption{\textbf{Command model MGSM results.} (Acc.)}
\label{tab:app-mgsm-raw}
\end{table*}

\begin{table*}[htb]
\footnotesize
\centering
\begin{tabular}{@{}llrrrrr|r|a|aa@{}}
\toprule
 &
   &
   &
   &
   &
   &
   &
   &
  \multicolumn{3}{c}{\textbf{non-en stats}} \\ \midrule
 &
   &
  \multicolumn{1}{c}{\textbf{de}} &
  \multicolumn{1}{c}{\textbf{es}} &
  \multicolumn{1}{c}{\textbf{fr}} &
  \multicolumn{1}{c}{\textbf{ja}} &
  \multicolumn{1}{c}{\textbf{zh}} &
  \multicolumn{1}{|c|}{\textbf{en}} &
  \multicolumn{1}{c}{\textbf{avg}} &
  \multicolumn{1}{c}{\textbf{Ltn/IE}} &
  \multicolumn{1}{c}{\textbf{$\neg$}} \\ \midrule
 &
  W8 &
  \cellcolor[HTML]{F7FCFA}0.3\% &
  \cellcolor[HTML]{FAE6E4}-0.9\% &
  \cellcolor[HTML]{F7D5D2}-1.6\% &
  \cellcolor[HTML]{F1B9B4}-2.7\% &
  \cellcolor[HTML]{FEFCFB}-0.1\% &
  \cellcolor[HTML]{F8FDFA}0.3\% &
  \cellcolor[HTML]{FAE4E3}-1.0\% &
  \cellcolor[HTML]{FBEBEA}-0.7\% &
  \cellcolor[HTML]{F8DAD8}-1.4\% \\
 &
  W8A8-sq &
  \cellcolor[HTML]{E3F4EC}1.1\% &
  \cellcolor[HTML]{EC9F98}-3.7\% &
  \cellcolor[HTML]{F7D8D5}-1.5\% &
  \cellcolor[HTML]{FEFBFB}-0.1\% &
  \cellcolor[HTML]{EC9F98}-3.6\% &
  \cellcolor[HTML]{EDA59E}-3.4\% &
  \cellcolor[HTML]{F6D4D1}-1.6\% &
  \cellcolor[HTML]{F8DBD9}-1.3\% &
  \cellcolor[HTML]{F5CDCA}-1.9\% \\
 &
  W8A8 &
  \cellcolor[HTML]{C9EADA}2.1\% &
  \cellcolor[HTML]{EDA29B}-3.5\% &
  \cellcolor[HTML]{F9E1DF}-1.1\% &
  \cellcolor[HTML]{EFF9F4}0.6\% &
  \cellcolor[HTML]{EFABA5}-3.2\% &
  \cellcolor[HTML]{FFFFFF}0.0\% &
  \cellcolor[HTML]{F9E4E2}-1.0\% &
  \cellcolor[HTML]{FAE8E6}-0.9\% &
  \cellcolor[HTML]{F8DDDB}-1.3\% \\
 &
  W4-g &
  \cellcolor[HTML]{F5CDCA}-1.9\% &
  \cellcolor[HTML]{F8DEDB}-1.3\% &
  \cellcolor[HTML]{F3C3BF}-2.3\% &
  \cellcolor[HTML]{E67C73}-7.9\% &
  \cellcolor[HTML]{F5CCC8}-1.9\% &
  \cellcolor[HTML]{E98C84}-4.4\% &
  \cellcolor[HTML]{EFAFA9}-3.0\% &
  \cellcolor[HTML]{F6CFCC}-1.8\% &
  \cellcolor[HTML]{E67E75}-4.9\% \\
\bottomrule 
\end{tabular}%
\caption{\textbf{Relative performance (\%$\Delta$) vs. FP16 for Command Models on MGSM}. Ltn/IE: Latin-script/Indo-European languages: de, es, fr.  $\neg$: ja, zh.
} 
\label{tab:app-mgsm-pct-drop}
\end{table*}

\begin{table*}[hp]
\footnotesize
\centering
\begin{tabular}{ll|cccccc|ca}
\toprule
 &
   &
  \textbf{de} &
  \textbf{es} &
  \textbf{fr} &
  \textbf{ja} &
  \textbf{ru} &
  \textbf{zh} &
  \textbf{en} & 
  \textbf{non-en Avg} \\ \midrule
           & FP16 & 61.6 & 58.4 & 55.6 & 22.8 & 58.0 & 50.8 & 68.4 & 51.2 \\
Aya-23-35b & W8   & 54.4 & 61.2 & 60.4 & 24.4 & 57.2 & 55.2 & 66.4 & 52.1 \\
           & W4   & 58.8 & 54.8 & 54.8 & 18.4 & 53.6 & 48.4 & 66.0 & 48.1 \\ \midrule
           & FP16 & 40.4 & 45.2 & 38.8 & 12.8 & 38.0 & 32.8 & 48.0 & 34.7 \\
Aya-23-8b  & W8   & 39.6 & 45.6 & 38.8 & 13.6 & 38.8 & 36.0 & 45.6 & 35.4 \\
           & W4   & 39.6 & 42.0 & 34.0 & 7.2  & 33.6 & 36.0 & 42.4 & 32.1 \\ \bottomrule
\end{tabular}%

\caption{\textbf{Aya 23 language-specific results for MGSM (5-shot).}}
\end{table*}

\begin{table*}[!htb]
\footnotesize
\centering
\begin{tabular}{@{}ll|rrrrrrr|cccc}
\toprule
 &
   &
  \multicolumn{1}{l}{} &
  \multicolumn{1}{l}{} &
  \multicolumn{1}{l}{} &
  \multicolumn{1}{l}{} &
  \multicolumn{1}{l}{} &
  \multicolumn{1}{l|}{} &
  \multicolumn{1}{l|}{} &
  \multicolumn{3}{c}{\textbf{non-en Stats}} \\ \midrule
 &
  \multicolumn{1}{l|}{} &
  \multicolumn{1}{c}{\textbf{de}} &
  \multicolumn{1}{c}{\textbf{es}} &
  \multicolumn{1}{c}{\textbf{fr}} &
  \multicolumn{1}{c}{\textbf{ja}} &
  \multicolumn{1}{c}{\textbf{ru}} &
  \multicolumn{1}{c|}{\textbf{zh}} &
  \multicolumn{1}{c|}{\textbf{en}} &
  \textbf{Avg} &
  \textbf{Ltn/IE} &
  \textbf{$\neg$} \\ \midrule
\multirow{2}{*}{Aya-23-35b} &
  \multicolumn{1}{l|}{W8} &
  \cellcolor[HTML]{EB9891}-11.7\% &
  \cellcolor[HTML]{ADDEC6}4.8\% &
  \cellcolor[HTML]{6BC398}8.6\% &
  \cellcolor[HTML]{87CFAB}7.0\% &
  \cellcolor[HTML]{FCF2F2}-1.4\% &
  \cellcolor[HTML]{6AC398}8.7\% &
  \cellcolor[HTML]{FAE5E3}-2.9\% &
  \cellcolor[HTML]{D1EDDF}2.7\% &
  \cellcolor[HTML]{FEFCFC}0.6\% &
  \cellcolor[HTML]{ADDEC6}4.8\% \\
 &
  \multicolumn{1}{l|}{W4} &
  \cellcolor[HTML]{F7D7D4}-4.5\% &
  \cellcolor[HTML]{F4C9C5}-6.2\% &
  \cellcolor[HTML]{FCF2F1}-1.4\% &
  \cellcolor[HTML]{E67C73}-19.3\% &
  \cellcolor[HTML]{F2BCB8}-7.6\% &
  \cellcolor[HTML]{F7D5D2}-4.7\% &
  \cellcolor[HTML]{F9E0DE}-3.5\% &
  \cellcolor[HTML]{F2BFBA}-7.3\% &
  \cellcolor[HTML]{F8DCDA}-4.0\% &
  \cellcolor[HTML]{EDA29C}-10.5\% \\
\multirow{2}{*}{Aya-23-8b} &
  \multicolumn{1}{l|}{W8} &
  \cellcolor[HTML]{FBEDEC}-2.0\% &
  \cellcolor[HTML]{F0F9F5}0.9\% &
  \cellcolor[HTML]{FFFFFF}0.0\% &
  \cellcolor[HTML]{94D4B5}6.2\% &
  \cellcolor[HTML]{DBF1E6}2.1\% &
  \cellcolor[HTML]{57BB8A}9.8\% &
  \cellcolor[HTML]{F6D3D0}-5.0\% &
  \cellcolor[HTML]{CFECDD}2.8\% &
  \cellcolor[HTML]{FCF1F0}-0.4\% &
  \cellcolor[HTML]{98D5B7}6.0\% \\
\bottomrule
\end{tabular}%
\caption{\textbf{Relative performance ($\%\Delta$) vs. FP16 for Aya 23 models on MGSM (5-shot).} Ltn/IE are non-English Latin-script/Indo-European languages: de, es, fr. $\neg$ are the rest: ja, ru, zh.}
\label{tab:app-aya-mgsm}
\end{table*}

\begin{table*}[htb]
\centering
\resizebox{\textwidth}{!}{%
\begin{tabular}{@{}ll|rrrrrrrrrrrrrrrrrrrrr|ra}
\toprule
 &  & \multicolumn{1}{c}{\textbf{ar}} & \multicolumn{1}{c}{\textbf{cs}} & \multicolumn{1}{c}{\textbf{de}} & \multicolumn{1}{c}{\textbf{el}} & \multicolumn{1}{c}{\textbf{es}} & \multicolumn{1}{c}{\textbf{fa}} & \multicolumn{1}{c}{\textbf{fr}} & \multicolumn{1}{c}{\textbf{hi}} & \multicolumn{1}{c}{\textbf{id}} & \multicolumn{1}{c}{\textbf{it}} & \multicolumn{1}{c}{\textbf{ja}} & \multicolumn{1}{c}{\textbf{ko}} & \multicolumn{1}{c}{\textbf{nl}} & \multicolumn{1}{c}{\textbf{pl}} & \multicolumn{1}{c}{\textbf{pt}} & \multicolumn{1}{c}{\textbf{ro}} & \multicolumn{1}{c}{\textbf{ru}} & \multicolumn{1}{c}{\textbf{tr}} & \multicolumn{1}{c}{\textbf{uk}} & \multicolumn{1}{c}{\textbf{vi}} & \multicolumn{1}{c}{\textbf{zh}} & \multicolumn{1}{|c}{\textbf{en}} & \textbf{non-en Avg} \\ \midrule
\multirow{3}{*}{35b} & FP16 & 78.9 & 78.2 & 77.1 & 76.4 & 81.0 & 75.8 & 81.9 & 65.6 & 77.8 & 79.8 & 75.9 & 73.3 & 77.7 & 75.8 & 83.8 & 78.9 & 79.6 & 74.1 & 77.6 & 78.3 & 81.2 & 84.7 & 77.6 \\
 & W8 & 77.3 & 78.8 & 77.2 & 76.6 & 80.8 & 74.9 & 82.4 & 65.6 & 77.6 & 80.8 & 74.8 & 73.7 & 77.6 & 74.8 & 82.9 & 77.1 & 78.9 & 72.0 & 77.2 & 77.0 & 80.3 & 84.6 & 77.1 \\
 & W4 & 73.8 & 74.9 & 73.2 & 70.8 & 77.0 & 71.4 & 78.1 & 61.0 & 73.9 & 76.2 & 71.7 & 67.4 & 73.0 & 70.4 & 80.1 & 74.3 & 73.3 & 68.2 & 73.0 & 71.4 & 78.8 & 83.2 & 73.0 \\ \midrule
\multirow{3}{*}{8b} & FP16 & 65.6 & 61.9 & 65.6 & 64.0 & 67.0 & 63.6 & 69.6 & 54.3 & 67.4 & 65.7 & 65.2 & 61.7 & 63.8 & 61.3 & 69.1 & 65.7 & 69.7 & 58.1 & 66.8 & 62.3 & 72.2 & 77.0 & 64.8 \\
 & W8 & 64.3 & 61.8 & 64.8 & 63.0 & 67.4 & 63.9 & 70.4 & 54.2 & 67.4 & 64.6 & 65.4 & 61.4 & 64.3 & 59.8 & 68.7 & 65.4 & 68.7 & 58.1 & 67.0 & 63.7 & 71.8 & 76.1 & 64.6 \\
 & W4 & 61.9 & 57.0 & 61.6 & 57.7 & 61.1 & 58.2 & 65.7 & 49.8 & 64.7 & 58.3 & 60.7 & 51.1 & 60.7 & 54.9 & 62.0 & 59.8 & 63.9 & 50.1 & 61.0 & 58.8 & 66.2 & 73.8 & 59.3 \\ \bottomrule
\end{tabular}%
}\caption{\textbf{Aya 23 language-specific results for Belebele.} (Accuracy)}
\end{table*}

\begin{table*}[htb]
\centering
\resizebox{\textwidth}{!}{%
\begin{tabular}{@{}ll|rrrrrrrrrrrrrrrrrrrrr|r|ccc}
\toprule
 &  & \multicolumn{1}{l}{} & \multicolumn{1}{l}{} & \multicolumn{1}{l}{} & \multicolumn{1}{l}{} & \multicolumn{1}{l}{} & \multicolumn{1}{l}{} & \multicolumn{1}{l}{} & \multicolumn{1}{l}{} & \multicolumn{1}{l}{} & \multicolumn{1}{l}{} & \multicolumn{1}{l}{} & \multicolumn{1}{l}{} & \multicolumn{1}{l}{} & \multicolumn{1}{l}{} & \multicolumn{1}{l}{} & \multicolumn{1}{l}{} & \multicolumn{1}{l}{} & \multicolumn{1}{l}{} & \multicolumn{1}{l}{} & \multicolumn{1}{l}{} & \multicolumn{1}{l|}{} & \multicolumn{1}{l|}{} & \multicolumn{3}{c}{\textbf{non-En Stats}} \\ \midrule
 &  & \multicolumn{1}{c}{\textbf{ar}} & \multicolumn{1}{c}{\textbf{cs}} & \multicolumn{1}{c}{\textbf{de}} & \multicolumn{1}{c}{\textbf{el}} & \multicolumn{1}{c}{\textbf{es}} & \multicolumn{1}{c}{\textbf{fa}} & \multicolumn{1}{c}{\textbf{fr}} & \multicolumn{1}{c}{\textbf{hi}} & \multicolumn{1}{c}{\textbf{id}} & \multicolumn{1}{c}{\textbf{it}} & \multicolumn{1}{c}{\textbf{ja}} & \multicolumn{1}{c}{\textbf{ko}} & \multicolumn{1}{c}{\textbf{nl}} & \multicolumn{1}{c}{\textbf{pl}} & \multicolumn{1}{c}{\textbf{pt}} & \multicolumn{1}{c}{\textbf{ro}} & \multicolumn{1}{c}{\textbf{ru}} & \multicolumn{1}{c}{\textbf{tr}} & \multicolumn{1}{c}{\textbf{uk}} & \multicolumn{1}{c}{\textbf{vi}} & \multicolumn{1}{c|}{\textbf{zh}} & \multicolumn{1}{c|}{\textbf{en}} & \textbf{Avg} & \textbf{Ltn} & \textbf{$\neg$} \\
\bottomrule
\end{tabular}%
}
\caption{\textbf{Relative performance ($\%\Delta$) vs. FP16 for Aya 23 models on Belebele.} Ltn are non-English Latin-script languages: cs, de, es, es, fr, id, it, nl, pl, pt, ro, tr, vi. $\neg$ are the rest.}
\end{table*}

\begin{table*}[htb]
\centering
\footnotesize
\begin{tabular}{@{}ll|rrrrrrrrr|ra@{}}
\toprule
 &
   &
  \multicolumn{1}{c}{\textbf{ar}} &
  \multicolumn{1}{c}{\textbf{de}} &
  \multicolumn{1}{c}{\textbf{es}} &
  \multicolumn{1}{c}{\textbf{fr}} &
  \multicolumn{1}{c}{\textbf{it}} &
  \multicolumn{1}{c}{\textbf{ja}} &
  \multicolumn{1}{c}{\textbf{ko}} &
  \multicolumn{1}{c}{\textbf{pt}} &
  \multicolumn{1}{c}{\textbf{zh}} &
  \multicolumn{1}{|c}{\textbf{en}} &
  \textbf{non-en Avg} \\ \midrule
\multirow{6}{*}{103B} & FP16    & 64.0 & 68.3 & 68.7 & 68.0 & 69.3 & 64.4 & 62.3 & 70.0 & 65.0 & 75.7 & 66.7 \\
                      & W8      & 64.1 & 68.3 & 68.7 & 68.1 & 69.4 & 64.3 & 62.3 & 69.9 & 65.0 & 75.7 & 66.7 \\
                      & W8A8-sq & 63.5 & 67.9 & 68.8 & 68.0 & 69.1 & 63.6 & 61.8 & 69.2 & 64.9 & 75.6 & 66.3 \\
                      & W8A8    & 62.6 & 67.1 & 68.2 & 67.4 & 68.3 & 62.9 & 60.8 & 68.7 & 64.1 & 75.2 & 65.6 \\
                      & W4-g    & 62.9 & 67.5 & 68.2 & 67.6 & 68.6 & 62.8 & 61.1 & 68.6 & 64.0 & 74.4 & 65.7 \\
                      & W4      & 60.5 & 65.7 & 66.5 & 65.4 & 66.6 & 61.1 & 59.3 & 66.7 & 62.1 & 73.2 & 63.8 \\ \midrule
\multirow{4}{*}{35B}  & FP16    & 56.5 & 60.7 & 62.3 & 61.8 & 62.0 & 56.4 & 54.8 & 62.0 & 57.9 & 67.7 & 59.4 \\
                      & W8      & 56.5 & 60.6 & 62.2 & 61.8 & 61.9 & 56.4 & 54.7 & 62.1 & 57.9 & 67.7 & 59.3 \\
                      & W8A8    & 56.4 & 60.5 & 62.5 & 61.9 & 62.0 & 55.8 & 54.5 & 61.8 & 58.1 & 67.7 & 59.3 \\
                      & W4-g    & 55.4 & 59.7 & 62.0 & 61.0 & 60.7 & 54.4 & 53.2 & 60.8 & 56.6 & 66.5 & 58.2 \\ \bottomrule
\end{tabular}%
\caption{\textbf{mMMLU scores for Command Models.} (Accuracy)}
\label{tab:app-m3lu-raw}
\end{table*}

\begin{table*}[htb]
\centering
\resizebox{\textwidth}{!}{%
\begin{tabular}{@{}ll|rrrrrrrrr||r|rrr@{}}
\toprule
                       & \multicolumn{1}{r}{} &                                 &                                 &                                 &                                 &                                 &                                 &                                 &                                 &                                 & \multicolumn{1}{l}{}            & \multicolumn{3}{|c}{\textbf{non-en Stats}}                                                                  \\ \midrule
                       &                      & \multicolumn{1}{c}{\textbf{ar}} & \multicolumn{1}{c}{\textbf{de}} & \multicolumn{1}{c}{\textbf{es}} & \multicolumn{1}{c}{\textbf{fr}} & \multicolumn{1}{c}{\textbf{it}} & \multicolumn{1}{c}{\textbf{ja}} & \multicolumn{1}{c}{\textbf{ko}} & \multicolumn{1}{c}{\textbf{pt}} & \multicolumn{1}{c||}{\textbf{zh}} & \multicolumn{1}{c}{\textbf{en}} & \multicolumn{1}{|c}{\textbf{Avg}} & \multicolumn{1}{c}{\textbf{Ltn/IE}} & \multicolumn{1}{c}{\textbf{$\neg$}} \\
                       & W8                   & \cellcolor[HTML]{FBFEFD}0.2\%   & \cellcolor[HTML]{FFFFFF}0.0\%   & \cellcolor[HTML]{FFFFFF}0.0\%   & \cellcolor[HTML]{FCFEFD}0.1\%   & \cellcolor[HTML]{FCFEFD}0.1\%   & \cellcolor[HTML]{FEFAFA}-0.2\%  & \cellcolor[HTML]{FFFFFF}0.0\%   & \cellcolor[HTML]{FEFBFA}-0.1\%  & \cellcolor[HTML]{FFFFFF}0.0\%   & \cellcolor[HTML]{FFFFFF}0.0\%   & \cellcolor[HTML]{FFFFFF}0.0\%    & \cellcolor[HTML]{FFFFFF}0.0\%       & \cellcolor[HTML]{FFFFFF}0.0\%     \\
\bottomrule
\end{tabular}%
}

\caption{\textbf{Relative performance (\%$\Delta$) vs. FP16 for Command Models on mMMLU}. Ltn/IE are non-English Latin-script/Indo-European languages: de, es, fr, it, pt.  $\neg$ are the rest: ar, ja, ko, zh.}
\label{tab:app-m3lu-pct-drop}
\end{table*}

\begin{table*}[htb]
\centering
\resizebox{\textwidth}{!}{%
\begin{tabular}{@{}cl|cccccccccccccc|ca}
\toprule
 &  & {\textbf{ar}} & {\textbf{de}} & {\textbf{es}} & {\textbf{fr}} & {\textbf{hi}} & {\textbf{id}} & {\textbf{it}} & {\textbf{nl}} & {\textbf{pt}} & {\textbf{ro}} & {\textbf{ru}} & {\textbf{uk}} & {\textbf{vi}} & \multicolumn{1}{c|}{\textbf{zh}} & {\textbf{en}} & {\textbf{non-en Avg}} \\ \midrule
 & FP16 & 53.9 & 60.4 & 61.6 & 62.0 & 47.8 & 58.9 & 61.5 & 60.3 & 62.0 & 59.7 & 57.8 & 56.3 & 55.3 & 57.5 & 66.7 & 58.2 \\
Aya-23-35b & W8 & 53.8 & 60.0 & 61.7 & 61.7 & 47.4 & 58.7 & 61.1 & 60.0 & 61.6 & 59.1 & 57.5 & 56.1 & 54.9 & 57.5 & 66.2 & 57.9 \\
 & W4 & 52.3 & 58.7 & 60.3 & 60.4 & 45.7 & 57.4 & 59.8 & 58.6 & 60.5 & 57.7 & 56.5 & 55.0 & 53.8 & 56.1 & 65.2 & 56.6 \\ \midrule
 & FP16 & 45.1 & 50.0 & 50.9 & 51.0 & 39.7 & 48.8 & 50.7 & 49.7 & 50.8 & 49.9 & 47.8 & 46.8 & 46.5 & 47.1 & 54.6 & 48.2 \\
Aya-23-8b & W8 & 44.9 & 49.9 & 50.5 & 50.6 & 39.4 & 48.5 & 50.2 & 49.4 & 50.6 & 49.2 & 47.4 & 46.3 & 45.7 & 46.4 & 54.2 & 47.8 \\
 & W4 & 43.9 & 48.4 & 49.4 & 49.0 & 38.4 & 47.5 & 49.1 & 47.9 & 49.1 & 48.0 & 46.2 & 45.6 & 44.9 & 46.1 & 53.4 & 46.7 \\ \bottomrule
\end{tabular}%
}\caption{\textbf{Aya 23 language-specific results for mMMLU (Okapi).} (Accuracy)}
\end{table*}

\begin{table*}[htb]
\centering
\resizebox{\textwidth}{!}{%
\begin{tabular}{@{}ll|rrrrrrrrrrrrrr|r|ccc}
 &  & & & & & & & & & & & & & & & & \multicolumn{3}{c}{\textbf{non-En Stats}} \\ \midrule
 &  & \multicolumn{1}{c}{\textbf{ar}} & \multicolumn{1}{c}{\textbf{de}} & \multicolumn{1}{c}{\textbf{es}} & \multicolumn{1}{c}{\textbf{fr}} & \multicolumn{1}{c}{\textbf{hi}} & \multicolumn{1}{c}{\textbf{id}} & \multicolumn{1}{c}{\textbf{it}} & \multicolumn{1}{c}{\textbf{nl}} & \multicolumn{1}{c}{\textbf{pt}} & \multicolumn{1}{c}{\textbf{ro}} & \multicolumn{1}{c}{\textbf{ru}} & \multicolumn{1}{c}{\textbf{uk}} & \multicolumn{1}{c}{\textbf{vi}} & \multicolumn{1}{c|}{\textbf{zh}} & \multicolumn{1}{c|}{\textbf{en}} & \textbf{Avg} & \textbf{Ltn} & \textbf{$\neg$} \\ \midrule
\multirow{2}{*}{Aya-23-35b} & W8 & \cellcolor[HTML]{FEFBFA}-0.2\% & \cellcolor[HTML]{FCF1F1}-0.7\% & \cellcolor[HTML]{F1F9F5}0.2\% & \cellcolor[HTML]{FDF7F6}-0.4\% & \cellcolor[HTML]{FCF2F1}-0.7\% & \cellcolor[HTML]{FDF9F8}-0.3\% & \cellcolor[HTML]{FDF4F3}-0.6\% & \cellcolor[HTML]{FDF6F5}-0.5\% & \cellcolor[HTML]{FCF4F3}-0.6\% & \cellcolor[HTML]{FBEBE9}-1.1\% & \cellcolor[HTML]{FCF3F2}-0.6\% & \cellcolor[HTML]{FDF9F8}-0.3\% & \cellcolor[HTML]{FCF0EF}-0.8\% & \cellcolor[HTML]{FEFCFC}-0.1\% & \cellcolor[HTML]{FCF3F2}-0.6\% & \cellcolor[HTML]{FDF6F5}-0.5\% & \cellcolor[HTML]{FDF5F4}-0.5\% & \cellcolor[HTML]{FDF7F7}-0.4\% \\
 & W4 & \cellcolor[HTML]{F4C5C1}-3.1\% & \cellcolor[HTML]{F5CBC7}-2.8\% & \cellcolor[HTML]{F7D8D6}-2.0\% & \cellcolor[HTML]{F6D0CD}-2.5\% & \cellcolor[HTML]{EFADA7}-4.4\% & \cellcolor[HTML]{F6D0CD}-2.5\% & \cellcolor[HTML]{F5CCC8}-2.7\% & \cellcolor[HTML]{F4C9C6}-2.8\% & \cellcolor[HTML]{F6D0CD}-2.5\% & \cellcolor[HTML]{F3C0BC}-3.3\% & \cellcolor[HTML]{F6D4D2}-2.2\% & \cellcolor[HTML]{F7D5D2}-2.2\% & \cellcolor[HTML]{F5CCC8}-2.7\% & \cellcolor[HTML]{F5CECB}-2.6\% & \cellcolor[HTML]{F7D6D3}-2.2\% & \cellcolor[HTML]{F5CCC9}-2.7\% & \cellcolor[HTML]{F5CECB}-2.6\% & \cellcolor[HTML]{F4C8C5}-2.9\% \\
\bottomrule
\end{tabular}%
}
\caption{\textbf{Relative performance ($\%\Delta$) vs. FP16 for Aya Models on mMMLU (Okapi).} Ltn are non-English Latin-script languages: de, es, fr, id, it, nl, pt, ro, vi. $\neg$ are the rest.}
\end{table*}

\begin{table*}[htb]
\centering
\resizebox{\textwidth}{!}{%
\begin{tabular}{@{}l|l|rrrrrrrrr|r||rrrrrrrrr|r@{}}
\toprule
 &
   &
  \multicolumn{10}{c|}{\textbf{English $\rightarrow$ L2}} &
  \multicolumn{10}{c}{\textbf{L2 $\rightarrow$ English}} \\ \midrule
\toprule
                      &         & \multicolumn{1}{c}{\textbf{ar}} & \multicolumn{1}{c}{\textbf{de}} & \multicolumn{1}{c}{\textbf{es}} & \multicolumn{1}{c}{\textbf{fr}} & \multicolumn{1}{c}{\textbf{it}} & \multicolumn{1}{c}{\textbf{ja}} & \multicolumn{1}{c}{\textbf{ko}} & \multicolumn{1}{c}{\textbf{pt}} & \multicolumn{1}{c|}{\textbf{zh}} & \multicolumn{1}{c|}{\textbf{Avg}} & \multicolumn{1}{c}{\textbf{ar}} & \multicolumn{1}{c}{\textbf{de}} & \multicolumn{1}{c}{\textbf{es}} & \multicolumn{1}{c}{\textbf{fr}} & \multicolumn{1}{c}{\textbf{it}} & \multicolumn{1}{c}{\textbf{ja}} & \multicolumn{1}{c}{\textbf{ko}} & \multicolumn{1}{c}{\textbf{pt}} & \multicolumn{1}{c|}{\textbf{zh}} & \multicolumn{1}{c}{\textbf{Avg}} \\ \midrule
\multirow{6}{*}{103B} & FP16    & 27.1                            & 40.0                            & 30.1                            & 50.6                            & 33.1                            & 33.1                            & 29.1                            & 51.0                            & 45.1                             & 37.7                              & 45.0                            & 46.3                            & 33.4                            & 48.6                            & 36.5                            & 29.5                            & 33.0                            & 52.2                            & 32.1                             & 39.6                             \\
\bottomrule
\end{tabular}%
}
\caption{\textbf{Full results on FLORES for Command Models.} (SacreBLEU)}
\label{tab:app-flores-raw}
\end{table*}

\begin{table*}[htb]
\centering
\resizebox{\textwidth}{!}{%
\begin{tabular}{@{}ll|rrrrrrrrr|rrr||rrrrrrrrr|rrr}
\toprule
 &
   &
  \multicolumn{12}{c|}{\textbf{English $\rightarrow$ L2}} &
  \multicolumn{12}{c}{\textbf{L2 $\rightarrow$ English}} \\ \midrule
\toprule
                       &         & \multicolumn{1}{c}{\textbf{ar}} & \multicolumn{1}{c}{\textbf{de}} & \multicolumn{1}{c}{\textbf{es}} & \multicolumn{1}{c}{\textbf{fr}} & \multicolumn{1}{c}{\textbf{it}} & \multicolumn{1}{c}{\textbf{ja}} & \multicolumn{1}{c}{\textbf{ko}} & \multicolumn{1}{c}{\textbf{pt}} & \multicolumn{1}{c|}{\textbf{zh}} & \multicolumn{1}{c}{\textbf{Avg}} & \multicolumn{1}{c}{\textbf{Ltn/IE}} & \multicolumn{1}{c|}{\textbf{$\neg$}} & \multicolumn{1}{|c}{\textbf{ar}} & \multicolumn{1}{c}{\textbf{de}} & \multicolumn{1}{c}{\textbf{es}} & \multicolumn{1}{c}{\textbf{fr}} & \multicolumn{1}{c}{\textbf{it}} & \multicolumn{1}{c}{\textbf{ja}} & \multicolumn{1}{c}{\textbf{ko}} & \multicolumn{1}{c}{\textbf{pt}} & \multicolumn{1}{c|}{\textbf{zh}} & \multicolumn{1}{c}{\textbf{Avg}} & \multicolumn{1}{c}{\textbf{Ltn/IE}} & \multicolumn{1}{c}{\textbf{$\neg$}} \\ \midrule
                       & W8      & \cellcolor[HTML]{FEFFFE}0.1\%   & \cellcolor[HTML]{FDFEFE}0.1\%   & \cellcolor[HTML]{FDF4F3}-0.4\%  & \cellcolor[HTML]{FBFEFD}0.2\%   & \cellcolor[HTML]{FCFEFD}0.1\%   & \cellcolor[HTML]{F8FCFA}0.3\%   & \cellcolor[HTML]{FDF9F9}-0.2\%  & \cellcolor[HTML]{FDF6F6}-0.3\%  & \cellcolor[HTML]{FCFEFD}0.1\%    & \cellcolor[HTML]{FEFEFE}0.0\%    & \cellcolor[HTML]{FEFDFD}-0.1\%      & \cellcolor[HTML]{FEFFFE}0.1\%      & \cellcolor[HTML]{F5FBF8}0.4\%   & \cellcolor[HTML]{FEFCFC}-0.1\%  & \cellcolor[HTML]{FEFBFB}-0.1\%  & \cellcolor[HTML]{FEFBFB}-0.1\%  & \cellcolor[HTML]{FEFDFD}-0.1\%  & \cellcolor[HTML]{FEFDFD}-0.1\%  & \cellcolor[HTML]{FFFFFF}0.0\%   & \cellcolor[HTML]{FEFCFC}-0.1\%  & \cellcolor[HTML]{FEFBFB}-0.1\%  & \cellcolor[HTML]{FEFEFE}0.0\%    & \cellcolor[HTML]{FEFCFC}-0.1\%      & \cellcolor[HTML]{FEFFFF}0.1\%     \\
\bottomrule
\end{tabular}%
}
\caption{\textbf{Relative performance (\%$\Delta$) vs. FP16 for Command Models on FLORES}. Ltn/IE are Latin-script/Indo-European languages: de, es, fr, it, pt.  $\neg$ are the rest: ar, ja, ko, zh.}
\label{tab:app-flores-pct-drop}
\end{table*}

\begin{table*}[htb]
\centering
\resizebox{\textwidth}{!}{%
\begin{tabular}{@{}ll|rrrrrrrrrrrrrrrrrrrrrr|a}
\toprule
 &  \multicolumn{24}{c}{\textbf{English$\rightarrow$L2}} \\ 
\textbf{}  &      & \multicolumn{1}{c}{\textbf{ar}} & \multicolumn{1}{c}{\textbf{cs}} & \multicolumn{1}{c}{\textbf{de}} & \multicolumn{1}{c}{\textbf{el}} & \multicolumn{1}{c}{\textbf{es}} & \multicolumn{1}{c}{\textbf{fa}} & \multicolumn{1}{c}{\textbf{fr}} & \multicolumn{1}{c}{\textbf{he}} & \multicolumn{1}{c}{\textbf{hi}} & \multicolumn{1}{c}{\textbf{id}} & \multicolumn{1}{c}{\textbf{it}} & \multicolumn{1}{c}{\textbf{ja}} & \multicolumn{1}{c}{\textbf{ko}} & \multicolumn{1}{c}{\textbf{nl}} & \multicolumn{1}{c}{\textbf{pl}} & \multicolumn{1}{c}{\textbf{pt}} & \multicolumn{1}{c}{\textbf{ro}} & \multicolumn{1}{c}{\textbf{ru}} & \multicolumn{1}{c}{\textbf{tr}} & \multicolumn{1}{c}{\textbf{uk}} & \multicolumn{1}{c}{\textbf{vi}} & \multicolumn{1}{c|}{\textbf{zh}} & \multicolumn{1}{c}{\textbf{Avg}} \\ \midrule
           
           & FP16 & 40.0                            & 39.1                            & 42.5                            & 36.3                            & 32.1                            & 33.4                            & 54.1                            & 39.5                            & 31.9                            & 44.7                            & 36.6                            & 28.7                            & 25.5                            & 33.4                            & 30.7                            & 53.1                            & 43.3                            & 38.9                            & 33.8                            & 38.2                            & 41.0                            & 34.0                             & 37.8                             \\
\bottomrule

\end{tabular}%
}
\caption{\textbf{Aya 23 language-specific results for FLORES.} spBLEU with FLORES200 tokenizer.}
\end{table*}

\begin{table*}[htb]
\centering
\resizebox{\textwidth}{!}{%
\begin{tabular}{ll|rrrrrrrrrrrrrrrrrrrrrr||rrr}
\toprule
 &  \multicolumn{24}{c}{\textbf{English$\rightarrow$L2}} \\ 
\textbf{}                    &    & \multicolumn{1}{c}{\textbf{ar}} & \multicolumn{1}{c}{\textbf{cs}} & \multicolumn{1}{c}{\textbf{de}} & \multicolumn{1}{c}{\textbf{el}} & \multicolumn{1}{c}{\textbf{es}} & \multicolumn{1}{c}{\textbf{fa}} & \multicolumn{1}{c}{\textbf{fr}} & \multicolumn{1}{c}{\textbf{he}} & \multicolumn{1}{c}{\textbf{hi}} & \multicolumn{1}{c}{\textbf{id}} & \multicolumn{1}{c}{\textbf{it}} & \multicolumn{1}{c}{\textbf{ja}} & \multicolumn{1}{c}{\textbf{ko}} & \multicolumn{1}{c}{\textbf{nl}} & \multicolumn{1}{c}{\textbf{pl}} & \multicolumn{1}{c}{\textbf{pt}} & \multicolumn{1}{c}{\textbf{ro}} & \multicolumn{1}{c}{\textbf{ru}} & \multicolumn{1}{c}{\textbf{tr}} & \multicolumn{1}{c}{\textbf{uk}} & \multicolumn{1}{c}{\textbf{vi}} & \multicolumn{1}{c|}{\textbf{zh}} & \multicolumn{1}{c}{\textbf{Avg}} & \multicolumn{1}{c}{\textbf{Ltn}} & \multicolumn{1}{c}{\textbf{$\neg$}} \\ \midrule
                             
                             & W8 & \cellcolor[HTML]{FEFDFD}-0.1\%  & \cellcolor[HTML]{FEFBFB}-0.2\%  & \cellcolor[HTML]{A6DBC2}1.1\%   & \cellcolor[HTML]{FEFDFD}-0.1\%  & \cellcolor[HTML]{F0F9F5}0.2\%   & \cellcolor[HTML]{C3E7D6}0.7\%   & \cellcolor[HTML]{FDF7F7}-0.4\%  & \cellcolor[HTML]{99D6B8}1.2\%   & \cellcolor[HTML]{91D3B2}1.3\%   & \cellcolor[HTML]{E3F4EC}0.3\%   & \cellcolor[HTML]{FDF8F7}-0.4\%  & \cellcolor[HTML]{D1EDDF}0.6\%   & \cellcolor[HTML]{FCFEFD}0.0\%   & \cellcolor[HTML]{CDEBDC}0.6\%   & \cellcolor[HTML]{C6E8D7}0.7\%   & \cellcolor[HTML]{E9F7F0}0.3\%   & \cellcolor[HTML]{DEF2E9}0.4\%   & \cellcolor[HTML]{FCF3F3}-0.6\%  & \cellcolor[HTML]{F6FBF9}0.1\%   & \cellcolor[HTML]{ECF7F2}0.2\%   & \cellcolor[HTML]{B2E0C9}0.9\%   & \cellcolor[HTML]{FCF0EF}-0.8\%   & \cellcolor[HTML]{E8F6EF}0.3\%    & \cellcolor[HTML]{E6F5EE}0.3\%    & \cellcolor[HTML]{EAF7F0}0.3\%     \\
\bottomrule
\end{tabular}%
}
\caption{\textbf{Relative performance ($\%\Delta$) vs. FP16 for Aya Models on FLORES.} Ltn are Latin-script languages: cs, de, es, fr, id, it, nl, pl, pt, ro, tr, vi. $\neg$ are the rest.}
\label{tab:app-aya-flores}
\end{table*}

\begin{table}[htb]
\centering
\resizebox{\textwidth}{!}{%
\begin{tabular}{@{}ll|rrr|rrr|rrr|rrr@{}}
\toprule
                             &                         & \multicolumn{3}{c}{\textbf{Avg Rel. \%Delta}}                                                    & \multicolumn{3}{|c}{\textbf{mMMLU}}                                                               & \multicolumn{3}{|c}{\textbf{MGSM}}                                                                  & \multicolumn{3}{|c}{\textbf{Belebele}}                                                            \\
                             &                         & \multicolumn{1}{c}{en}         & \multicolumn{1}{c}{Ltn}        & \multicolumn{1}{c}{All}        & \multicolumn{1}{|c}{en}         & \multicolumn{1}{c}{Ltn}        & \multicolumn{1}{c}{All}        & \multicolumn{1}{|c}{en}          & \multicolumn{1}{c}{Ltn}        & \multicolumn{1}{c}{All}         & \multicolumn{1}{|c}{en}         & \multicolumn{1}{c}{Ltn}        & \multicolumn{1}{c}{All}        \\ \midrule
                             & \multicolumn{1}{l|}{W8} & \cellcolor[HTML]{F8DEDC}-1.2\% & \cellcolor[HTML]{FEFAFA}-0.2\% & \cellcolor[HTML]{F2FAF6}0.5\%  & \cellcolor[HTML]{FBEEED}-0.6\% & \cellcolor[HTML]{FCF1F0}-0.5\% & \cellcolor[HTML]{FCF2F1}-0.5\% & \cellcolor[HTML]{F0B2AD}-2.9\%  & \cellcolor[HTML]{F1F9F5}0.6\%  & \cellcolor[HTML]{BAE4CF}2.7\%   & \cellcolor[HTML]{FEFBFB}-0.1\% & \cellcolor[HTML]{FCEFEE}-0.6\% & \cellcolor[HTML]{FBEEED}-0.6\% \\
\multirow{-2}{*}{Aya-23-35b} & \multicolumn{1}{l|}{W4} & \cellcolor[HTML]{F2BEB9}-2.5\% & \cellcolor[HTML]{EA928B}-4.1\% & \cellcolor[HTML]{E67C73}-5.3\% & \cellcolor[HTML]{F4C5C1}-2.2\% & \cellcolor[HTML]{F1B9B4}-2.6\% & \cellcolor[HTML]{F1B7B2}-2.7\% & \cellcolor[HTML]{EDA39C}-3.5\%  & \cellcolor[HTML]{EA948D}-4.0\% & \cellcolor[HTML]{E67C73}-7.3\%  & \cellcolor[HTML]{F6D2CF}-1.7\% & \cellcolor[HTML]{E67C73}-5.7\% & \cellcolor[HTML]{E67C73}-6.0\% \\
\bottomrule
\end{tabular}%
}
\caption{\textbf{Relative performance of quantized Aya 23 models in English vs. other languages.} All non-English languages (All), non-English Latin-script languages (Ltn).}
\label{tab:app-results-vs-en-aya}
\end{table}

\begin{table}[htb]
\footnotesize
\centering
\begin{tabular}{@{}l|l|cccc@{}}
\toprule
 &      & \multicolumn{1}{l}{Avg} & \multicolumn{1}{l}{XSC} & \multicolumn{1}{l}{XCOPA}    & \multicolumn{1}{l}{XWNG} \\ \midrule
 
                               & FP16 & 70.8                    & 65.1                              & 62.8                         & 84.4                          \\
Aya-23-35b                     & W8   & 70.6                    & 65.0                              & 62.9                         & 83.9                          \\
                               & W4   & 70.5                    & \cellcolor[HTML]{FFFFFF}64.8      & \cellcolor[HTML]{FFFFFF}62.3 & \cellcolor[HTML]{FFFFFF}84.5  \\ \midrule

                              & FP16 & 67.6                    & 62.3                              & 59.8                         & 80.7                          \\
Aya-23-8b                      & W8   & 67.6                    & 62.4                              & 60.0                         & 80.6                          \\
                               & W4   & 67.5                    & 62.3                              & 59.6                         & 80.6                          \\
\bottomrule
\end{tabular}
\caption{\textbf{Performance of quantized Aya 23 models on unseen discriminative tasks}. XStoryCloze (XSC), XCOPA, and XWinograd (XWNG).}
\label{tab:app-aya-unseen}
\end{table}

\begin{table*}[htb]
\centering
\resizebox{\textwidth}{!}{%
\begin{tabular}{@{}ll|rrrrrrrrr|ra||rrrrrrrrra}
\toprule
&  & \multicolumn{11}{c||}{\textbf{Monolingual}} & \multicolumn{10}{c}{\textbf{Cross-Lingual}} \\
 &  & \multicolumn{1}{c}{\textbf{ar}} & \multicolumn{1}{c}{\textbf{de}} & \multicolumn{1}{c}{\textbf{es}} & \multicolumn{1}{c}{\textbf{fr}} & \multicolumn{1}{c}{\textbf{it}} & \multicolumn{1}{c}{\textbf{ja}} & \multicolumn{1}{c}{\textbf{ko}} & \multicolumn{1}{c}{\textbf{pt}} & \multicolumn{1}{c|}{\textbf{zh}} & \multicolumn{1}{c}{\textbf{en}} & \textbf{Avg} & \multicolumn{1}{c}{\textbf{ar}} & \multicolumn{1}{c}{\textbf{de}} & \multicolumn{1}{c}{\textbf{es}} & \multicolumn{1}{c}{\textbf{fr}} & \multicolumn{1}{c}{\textbf{it}} & \multicolumn{1}{c}{\textbf{ja}} & \multicolumn{1}{c}{\textbf{ko}} & \multicolumn{1}{c}{\textbf{pt}} & \multicolumn{1}{c}{\textbf{zh}} & \multicolumn{1}{c}{\textbf{Avg}} \\ \midrule
\multirow{6}{*}{103B} & FP16 & 99.3 & 100.0 & 99.3 & 99.6 & 100.0 & 98.6 & 100.0 & 98.3 & 97.9 & 100.0 & 99.2 & 93.0 & 90.6 & 91.2 & 91.6 & 93.0 & 93.1 & 91.1 & 88.3 & 91.3 & 91.5 \\
 & W8 & 99.0 & 100.0 & 99.5 & 99.4 & 99.8 & 99.2 & 99.8 & 97.8 & 98.5 & 100.0 & 99.2 & 92.6 & 91.1 & 91.7 & 91.4 & 92.9 & 92.8 & 91.3 & 87.4 & 89.7 & 91.2 \\
 & W8A8-sq & 99.4 & 100.0 & 99.3 & 99.6 & 100.0 & 98.6 & 100.0 & 97.7 & 98.4 & 100.0 & 99.2 & 93.3 & 91.5 & 91.4 & 92.4 & 92.1 & 93.3 & 92.1 & 87.6 & 90.0 & 91.5 \\
 & W8A8 & 99.3 & 100.0 & 99.5 & 99.8 & 100.0 & 99.0 & 99.8 & 98.1 & 99.1 & 99.9 & 99.4 & 91.3 & 89.3 & 91.0 & 91.1 & 91.8 & 93.0 & 89.3 & 87.3 & 89.2 & 90.4 \\
 & W4-g & 99.1 & 100.0 & 99.6 & 99.9 & 100.0 & 97.4 & 100.0 & 98.1 & 98.9 & 100.0 & 99.2 & 90.6 & 89.9 & 90.7 & 91.7 & 93.1 & 92.8 & 90.6 & 85.4 & 89.6 & 90.5 \\
 & W4 & 99.4 & 100.0 & 99.4 & 99.7 & 99.8 & 99.6 & 99.0 & 98.9 & 98.4 & 100.0 & 99.3 & 95.8 & 94.3 & 95.9 & 93.8 & 93.6 & 92.6 & 88.9 & 90.5 & 89.7 & 92.8 \\
\bottomrule
\end{tabular}%
}
\caption{\textbf{Language Confusion scores for Command Models.} (Line-level pass rate (LPR))}
\label{tab:app-langconf-raw}
\end{table*}

\begin{table*}[htb]
\centering
\resizebox{\textwidth}{!}{%
\begin{tabular}{@{}ll|rrrrrrrrr|r|arr}
\toprule
 &  & \multicolumn{13}{c}{\textbf{Monolingual}} \\ 
 &  & \multicolumn{1}{c}{\textbf{ar}} & \multicolumn{1}{c}{\textbf{de}} & \multicolumn{1}{c}{\textbf{es}} & \multicolumn{1}{c}{\textbf{fr}} & \multicolumn{1}{c}{\textbf{it}} & \multicolumn{1}{c}{\textbf{ja}} & \multicolumn{1}{c}{\textbf{ko}} & \multicolumn{1}{c}{\textbf{pt}} & \multicolumn{1}{c|}{\textbf{zh}} & \multicolumn{1}{c|}{\textbf{en}} & \textbf{Avg} & \multicolumn{1}{c}{\textbf{Ltn/IE}} & \multicolumn{1}{c}{\textbf{$\neg$}} \\ \midrule
 & W8 & \cellcolor[HTML]{FDF7F7}-0.3\% & \cellcolor[HTML]{FFFFFF}0.0\% & \cellcolor[HTML]{FAFDFC}0.2\% & \cellcolor[HTML]{FEFAF9}-0.2\% & \cellcolor[HTML]{FDF9F9}-0.2\% & \cellcolor[HTML]{F0F9F5}0.6\% & \cellcolor[HTML]{FDF9F9}-0.2\% & \cellcolor[HTML]{FCF1F0}-0.5\% & \cellcolor[HTML]{F0F9F4}0.6\% & \multicolumn{1}{r|}{\cellcolor[HTML]{FFFFFF}0.0\%} & \cellcolor[HTML]{FFFFFF}0.0\% & \cellcolor[HTML]{FEFBFB}-0.1\% & \cellcolor[HTML]{FBFEFC}0.2\% \\
 & W8A8-sq & \cellcolor[HTML]{FEFFFE}0.1\% & \cellcolor[HTML]{FFFFFF}0.0\% & \cellcolor[HTML]{FFFFFF}0.0\% & \cellcolor[HTML]{FEFFFF}0.0\% & \cellcolor[HTML]{FFFFFF}0.0\% & \cellcolor[HTML]{FFFFFF}0.0\% & \cellcolor[HTML]{FFFFFF}0.0\% & \cellcolor[HTML]{FBEFED}-0.6\% & \cellcolor[HTML]{F2FAF6}0.5\% & \multicolumn{1}{r|}{\cellcolor[HTML]{FFFFFF}0.0\%} & \cellcolor[HTML]{FFFFFF}0.0\% & \cellcolor[HTML]{FEFCFB}-0.1\% & \cellcolor[HTML]{FCFEFD}0.1\% \\
 & W8A8 & \cellcolor[HTML]{FEFEFE}0.0\% & \cellcolor[HTML]{FFFFFF}0.0\% & \cellcolor[HTML]{FAFDFC}0.2\% & \cellcolor[HTML]{FAFDFC}0.2\% & \cellcolor[HTML]{FFFFFF}0.0\% & \cellcolor[HTML]{F5FBF8}0.4\% & \cellcolor[HTML]{FDF9F9}-0.2\% & \cellcolor[HTML]{FDF9F9}-0.2\% & \cellcolor[HTML]{E0F3E9}1.2\% & \multicolumn{1}{r|}{\cellcolor[HTML]{FEFCFC}-0.1\%} & \cellcolor[HTML]{FBFEFC}0.2\% & \cellcolor[HTML]{FEFFFF}0.0\% & \cellcolor[HTML]{F6FCF9}0.4\% \\
\bottomrule
\end{tabular}%
}

\caption{\textbf{Relative performance (\%$\Delta$) vs. FP16 for Command Models on Language Confusion metrics}. Ltn/IE are non-English Latin-script/Indo-European languages: de, es, fr, it, pt.  $\neg$ are the rest: ar, ja, ko, zh.}
\label{tab:app-langconf-pct}
\end{table*}

\clearpage
\newpage

\subsection{RM/LLM-as-a-Judge and Human Evaluation - Full Results}

\begin{table}[htb]
\footnotesize
\centering
\begin{tabular}{@{}ll|rr|rr|rr|rr}
\toprule
 &
   &
  \multicolumn{2}{c|}{\textbf{fr}} &
  \multicolumn{2}{c|}{\textbf{es}} &
  \multicolumn{2}{c|}{\textbf{ja}} &
  \multicolumn{2}{c}{\textbf{ko}}  \\
\textbf{} &
  \multicolumn{1}{l|}{\textbf{}} &
  \multicolumn{1}{c}{\textbf{LLM}} &
  \multicolumn{1}{c|}{\textbf{RM}} &
  \multicolumn{1}{c}{\textbf{LLM}} &
  \multicolumn{1}{c|}{\textbf{RM}} &
  \multicolumn{1}{c}{\textbf{LLM}} &
  \multicolumn{1}{c|}{\textbf{RM}} &
  \multicolumn{1}{c}{\textbf{LLM}} &
  \multicolumn{1}{c}{\textbf{RM}}  \\ \midrule
 &
  W8 &
  \cellcolor[HTML]{F3FAF7}50.5 &
  \cellcolor[HTML]{FEFCFC}49.7 &
  \cellcolor[HTML]{F7D6D3}44.9 &
  \cellcolor[HTML]{A1D9BE}53.7 &
  \cellcolor[HTML]{FAE9E8}47.3 &
  \cellcolor[HTML]{BBE4D0}52.7 &
  \cellcolor[HTML]{A2D9BE}53.7 &
  \cellcolor[HTML]{FAE8E6}47.1 \\
 &
  W8A8-sq &
  \cellcolor[HTML]{F1B6B1}40.8 &
  \cellcolor[HTML]{FBEAE9}47.5 &
  \cellcolor[HTML]{FCF0EF}48.1 &
  \cellcolor[HTML]{CCEBDC}52.0 &
  \cellcolor[HTML]{E6F5EE}51.0 &
  \cellcolor[HTML]{C4E7D6}52.4 &
  \cellcolor[HTML]{D0ECDF}51.9 &
  \cellcolor[HTML]{FBEAE9}47.5 \\
 &
  W4-g &
  \cellcolor[HTML]{F7D5D2}44.8 &
  \cellcolor[HTML]{F2BBB6}41.5 &
  \cellcolor[HTML]{F2BDB8}41.7 &
  \cellcolor[HTML]{E6F5EE}51.0 &
  \cellcolor[HTML]{F3C2BE}42.4 &
  \cellcolor[HTML]{FFFFFF}50.0 &
  \cellcolor[HTML]{FAE8E6}47.1 &
  \cellcolor[HTML]{F3C0BC}42.2 \\
\multirow{-4}{*}{Internal} &
  W4 &
  \cellcolor[HTML]{E8877F}34.9 &
  \cellcolor[HTML]{EFAEA8}39.8 &
  \cellcolor[HTML]{E67C73}33.5 &
  \cellcolor[HTML]{F2BBB6}41.5 &
  \cellcolor[HTML]{EEA8A2}39.2 &
  \cellcolor[HTML]{EFAFAA}40.0 &
  \cellcolor[HTML]{F0B5AF}40.7 &
  \cellcolor[HTML]{EA918A}36.2 \\ \midrule\midrule
 &
  W8 &
  \cellcolor[HTML]{FDF9F9}49.3 &
  \cellcolor[HTML]{E6F5EE}51.0 &
  \cellcolor[HTML]{A3DABF}53.7 &
  \cellcolor[HTML]{FBEFEE}48.0 &
  \cellcolor[HTML]{FAE7E5}47.0 &
  \cellcolor[HTML]{FAE9E8}47.3 &
  \cellcolor[HTML]{DEF2E8}51.3 &
  \cellcolor[HTML]{E6F5EE}51.0 \\
\bottomrule 
\end{tabular}
\caption{\textbf{\textit{LLM/RM-as-a-Judge} Raw win-rates of 103B quantized models vs. FP16} over \textit{Internal} and \textit{Aya Dolly} subsampled test sets.}
\label{tab:app-llm-as-a-judge-raw}
\end{table}

\begin{table}[htb]
\footnotesize
\centering
\begin{tabular}{@{}ll|rrrr||r|rcr}
\toprule
 &  & \multicolumn{1}{l}{} & \multicolumn{1}{l}{} & \multicolumn{1}{l}{} & \multicolumn{1}{l||}{} & \multicolumn{1}{l|}{} & \multicolumn{3}{c}{non-English Stats} \\ \midrule
 &  & \multicolumn{1}{c}{fr} & \multicolumn{1}{c}{es} & \multicolumn{1}{c}{ja} & \multicolumn{1}{c||}{ko} & \multicolumn{1}{c|}{en} & \multicolumn{1}{c}{avg} & \multicolumn{1}{c}{Ltn/IE} & \multicolumn{1}{c}{$\neg$} \\ \midrule
 & W8 & \cellcolor[HTML]{F3C4C0}46.3 & \cellcolor[HTML]{F7FCFA}50.3 & \cellcolor[HTML]{98D6B7}53.7 & \cellcolor[HTML]{ECA099}44.0 & \cellcolor[HTML]{F8DFDD}48.0 & \cellcolor[HTML]{FAE8E6}48.6 & \cellcolor[HTML]{F9E4E2}48.3 & \cellcolor[HTML]{FBECEB}48.9 \\
 & W8A8-sq & \cellcolor[HTML]{F0B4AF}45.3 & \cellcolor[HTML]{F3C4C0}46.3 & \cellcolor[HTML]{FBEFEE}49.0 & \cellcolor[HTML]{C8E9D9}52.0 & \cellcolor[HTML]{A3DABF}53.3 & \cellcolor[HTML]{F9E1DF}48.2 & \cellcolor[HTML]{F2BCB8}45.8 & \cellcolor[HTML]{F2FAF6}50.5 \\
\multirow{-3}{*}{Internal} & W4-g & \cellcolor[HTML]{E67C73}41.7 & \cellcolor[HTML]{F8DAD8}47.7 & \cellcolor[HTML]{E68078}42.0 & \cellcolor[HTML]{F8DAD8}47.7 & \cellcolor[HTML]{F3C4C0}46.3 & \cellcolor[HTML]{EFACA6}44.8 & \cellcolor[HTML]{EFABA5}44.7 & \cellcolor[HTML]{EFADA8}44.9 \\ \midrule
 & W8 & \cellcolor[HTML]{F7FCFA}50.3 & \cellcolor[HTML]{F6D4D1}47.3 & \cellcolor[HTML]{57BB8A}56.0 & \cellcolor[HTML]{FFFFFF}50.0 & \cellcolor[HTML]{F5CFCC}47.0 & \cellcolor[HTML]{E6F5EE}50.9 & \cellcolor[HTML]{FBECEA}48.8 & \cellcolor[HTML]{ABDDC5}53.0 \\
 & W8A8-sq & \cellcolor[HTML]{F3C4C0}46.3 & \cellcolor[HTML]{F2BBB6}45.7 & \cellcolor[HTML]{FFFFFF}50.0 & \cellcolor[HTML]{F9E4E2}48.3 & \cellcolor[HTML]{E3F4EC}51.0 & \cellcolor[HTML]{F7D8D6}47.6 & \cellcolor[HTML]{F2BFBB}46.0 & \cellcolor[HTML]{FCF1F0}49.2 \\
\bottomrule
\end{tabular}%
\caption{\textbf{Human evaluation raw win-rates of 103B quantized models vs.~FP16} over \emph{Internal} and \emph{Aya Dolly} subsampled test sets.}
\label{tab:human-eval-raw}
\end{table}

\end{document}